\documentclass[lettersize,journal]{IEEEtran}
\usepackage{amsmath,amssymb,graphicx}
\usepackage[dvipsnames]{xcolor}
\usepackage{bm,amsfonts}
\usepackage[pagebackref,breaklinks,colorlinks,bookmarks=false]{hyperref}
\usepackage{nicefrac}
\usepackage[switch]{lineno}
\usepackage{acronym}
\usepackage{resizegather}
\usepackage{booktabs}
\usepackage{pifont}
\usepackage[ruled,vlined,linesnumbered]{algorithm2e}    

\SetCommentSty{mycommfont}

\acrodef{CRF}{camera response function}
\acrodef{EXIF}{exchangeable image file format}
\acrodef{HDR}{high dynamic range}
\acrodef{MLE}{maximum likelihood estimation}
\acrodef{RANSAC}{random sample consensus}
\acrodef{BTF}{brightness transfer function}
\acrodef{MST}{maximum spanning trees}
\acrodef{RMSE}{root-mean-squared error}
\newcommand{\ourtitle}{Robust estimation of exposure ratios in multi-exposure image stacks}

\newcommand{\figref}[1]{Figure~\ref{fig:#1}}
\newcommand{\secref}[1]{Section~\ref{sec:#1}}

\newcommand{\tableref}[1]{Table~\ref{tab:#1}}
\newcommand{\algref}[1]{Algorithm~\ref{algo:#1}}

\LetLtxMacro{\originaleqref}{\eqref}
\renewcommand{\eqref}[1]{Eq.~\originaleqref{eq:#1}}

\newcommand{\E}{\mathbb{E}}
\newcommand{\V}{\mathbb{V}}
\newcommand{\transp}{^\mathsf{T}}

\DeclareMathOperator*{\argmin}{arg\,min}
\newcommand\edge{\mathrel{\bullet\mkern-3mu{-}\mkern-3mu\bullet}}

\newcommand{\rev}[1]{#1}
\newcommand{\tci}[1]{#1}
\newcommand{\tcirev}[1]{#1}

\usepackage{amsmath,amsfonts}
\usepackage{algorithmic}
\usepackage{array}
\usepackage[caption=false,font=normalsize,labelfont=sf,textfont=sf]{subfig}
\usepackage{textcomp}
\usepackage{stfloats}
\usepackage{url}
\usepackage{verbatim}
\usepackage{graphicx}
\usepackage{orcidlink}
\usepackage{cite}
\begin{document}

\title{\ourtitle}
% \author{Param Hanji\,\orcidlink{0000-0002-7985-4177}     
%     \and
%     Rafa{\l} K. Mantiuk\,\orcidlink{0000-0003-2353-0349}}

\author{\IEEEauthorblockN{Param Hanji\IEEEauthorrefmark{1},
Rafa{\l} K. Mantiuk\IEEEauthorrefmark{2}}
\IEEEauthorblockA{\\Department of Computer Science and Technology, 
University of Cambridge\\
Email: \{\IEEEauthorrefmark{1}pmh64,\IEEEauthorrefmark{2}rkm38\}@cam.ac.uk}}

% \markboth{Journal of \LaTeX\ Class Files,~Vol.~X, No.~X, XXX~2023}%
% {Hanji and Mantiuk: \ourtitle}


\maketitle

\let\thefootnote\relax\footnotetext{This work is published in the Transactions of Computational Imaging, 2023.

\copyright \, 2023 IEEE. Personal use of this material is permitted. Permission from IEEE must be obtained for all other uses, in any current or future media, including reprinting/republishing this material for advertising or promotional purposes, creating new collective works, for resale or redistribution to servers or lists, or reuse of any copyrighted component of this work in other works.

Digital Object Identifier 10.1109/TCI.2023.3301338}

\begin{abstract}
Merging multi-exposure image stacks into a high dynamic range (HDR) image requires knowledge of accurate exposure times. When exposure times are inaccurate, for example, when they are extracted from a camera's EXIF metadata, the reconstructed HDR images reveal banding artifacts at smooth gradients. To remedy this, we propose to estimate exposure ratios directly from the input images. We derive the exposure time estimation as an optimization problem, in which pixels are selected from pairs of exposures to minimize estimation error caused by camera noise. When pixel values are represented in the logarithmic domain, the problem can be solved efficiently using a linear solver. We demonstrate that the estimation can be easily made robust to pixel misalignment caused by camera or object motion by collecting pixels from multiple spatial tiles. The proposed automatic exposure estimation and alignment eliminates banding artifacts in popular datasets and is essential for applications that require physically accurate reconstructions, such as measuring the modulation transfer function of a display. The code for the method is available.
\end{abstract}


% \IEEEraisesectionheading{
%   \section{Introduction} \label{sec:introduction}
% }

\section{Introduction}

\IEEEPARstart{W}{hen} the dynamic range of a scene exceeds the operating range of a standard digital sensor, one can overcome this limitation by capturing a stack of images with different camera settings: modulating exposure times \cite{Gallo2016}, sensor gain  \cite{hajisharif2015adaptive,heide2014flexisp} or by capturing and averaging a burst of images \cite{hasinoff2016burst}. Then, the captured exposure stack can be merged into a single image to both expand the dynamic range and reduce noise. 

\begin{figure}
    \centering
    \includegraphics[width=0.32\columnwidth]{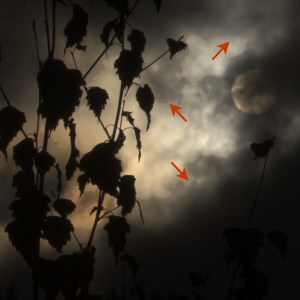}
    \includegraphics[width=0.32\columnwidth]{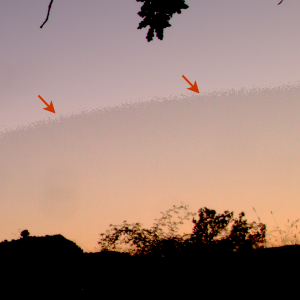}
    \includegraphics[width=0.32\columnwidth]{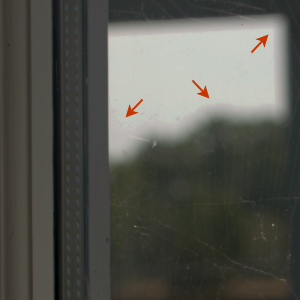}\\[0.5ex]
    \includegraphics[width=0.32\columnwidth]{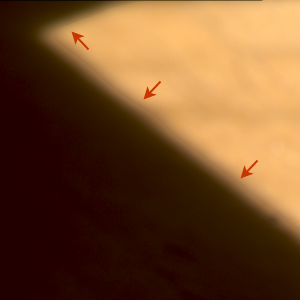}
    \includegraphics[width=0.32\columnwidth]{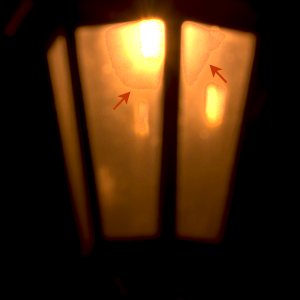}
    \includegraphics[width=0.32\columnwidth]{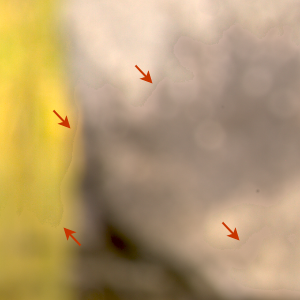}
    \caption{The first row depicts patches from images of overcast skies taken from large \ac{HDR} datasets \cite{fairchild2007hdr,hanji2022sihdr}. Misalignment between the reported camera settings results in banding artifacts (pointed at by the red arrows) at the boundary between two exposures. This problem is not limited to the sky and affects other regions too, as shown by patches in the second row. \rev{We carefully picked exposure stacks consisting of RAW spatially-aligned images that do not require \ac{CRF} inversion or motion compensation.}}
    \label{fig:problem}
\end{figure}

Regardless of the approach, regions of the reconstructed \ac{HDR} images that contain smooth intensity gradients often end up with banding artifacts, such as those shown in \figref{problem}. \tci{Banding artifacts are highly visible in both images and videos, and they have been the subject of extensive research, particularly in the field of video streaming. Banding tends to be more noticeable compared to other video compression artifacts~\cite{wang2016banding,tu2020banding,Kim2020}.} The reason for banding in merged exposure stacks is a mismatch between the actual capture parameters and those reported by the camera, typically in the \ac{EXIF} header. Such inaccuracies could be caused by
\begin{itemize}
    \item limited accuracy of the (mechanical) aperture and shutter
    % \PH{One of the reviewers wrote \emph{"most modern cameras come with extremely precise electronic shutters which are accurate to microseconds."} Should we remove this bullet?} \RM{Not, it stays. Many cameras still use analog shutters and the statement about the apperture is correct.}
    \item wrongly reported \ac{EXIF} data due to rounding (e.g. \nicefrac{1}{60} exposure time could be reported instead of \nicefrac{1}{64})
    \item changes in scene illumination, for example, due to flickering lights or overcast skies with moving clouds
\end{itemize}
\rev{We stress that the artifacts shown in \figref{problem} are not due to spatial misalignment because we carefully selected image stacks without camera or object motion. Further, since we merged linear RAW images, the
artifacts are not due to incorrect \ac{CRF} inversion.}

\begin{figure}
    \centering
    \includegraphics[width=\columnwidth]{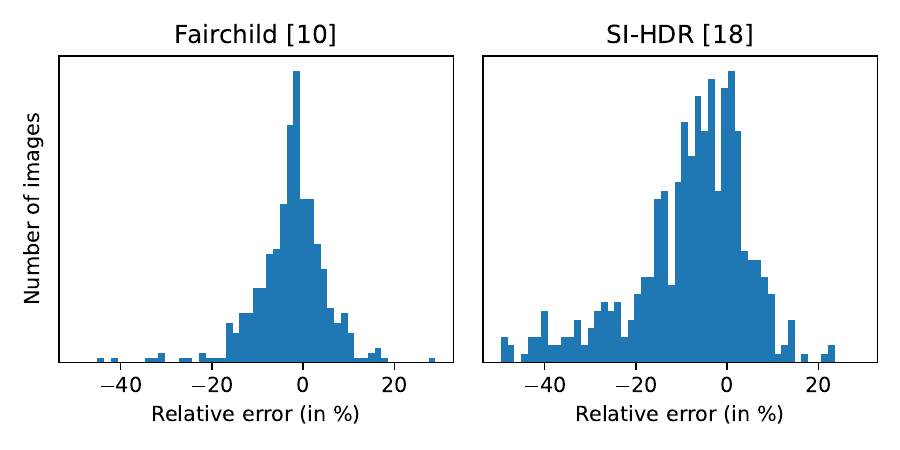}
    \caption{\rev{Histograms of relative errors for two popular HDR datasets. The computed sample standard deviations of relative errors are $9\%$ for the Fairchild survey~\cite{fairchild2007hdr} and $14\%$ for the SI-HDR dataset~\cite{hanji2022sihdr}.}}
    \label{fig:error-ratios}
\end{figure}

\rev{The inaccuracies of exposure time found in the EXIF metadata are substantial, reaching 40\%, as shown in \figref{error-ratios}. To prepare this figure, we computed the relative error between the reported EXIF exposure times and the times estimated by the proposed method. The two plots show similar error distribution for two datasets of multi-exposure HDR image stacks \cite{fairchild2007hdr,hanji2022sihdr}. The errors of such magnitude can easily introduce banding artifacts, as shown in \figref{problem}.}

% Additionally, the errors are randomly distributed around 0, i.e., they are unbiased and can not be accounted for by a simple calibration procedure.}

In this work, we propose to estimate exposure ratios directly from a stack of images, which lets us eliminate the banding artifacts from the merged HDR images. We formulate the problem to be efficiently solved as a sparse linear system while accounting for heteroskedastic camera noise with known or unknown noise parameters. Thus, our exposure estimation can be used to improve ground truth reconstructions of several existing stack-based \ac{HDR} datasets. \rev{Further, algorithms that enhance merged reconstructions by iterative optimization~\cite{ma2017mefssim}, as well as deep neural networks that use \ac{HDR} images for other related tasks~\cite{yang2018image,chen2020learning} stand to be improved by our more accurate exposures.}

\rev{Correctly estimating exposures} is also relevant for HDR merging with deghosting, which involves detecting and registering pixels that belong to moving objects \cite{tursun2015state,Perez-Pellitero_2021_CVPR}. Erroneous capture parameters make this task more challenging, as pixel differences could be caused by either object movement or incorrect exposure time.

\rev{Although banding artifacts are visible only in smooth image regions in \figref{problem}, inaccurate exposures affect all pixels in the image. Incorrect pixel intensities pose a serious complication for applications that utilize cameras in place of expensive light measurement instruments.} Multi-image exposure stacks often serve as substitutes for more accurate measurements from instruments such as spectrophotometers. In \secref{display-mtf}, we highlight potential discrepancies in the measured modulation transfer function (MTF) of an \ac{HDR} display when the exposure stack of a slant-edge~\cite{samei1998slantedge} is merged with incorrectly reported capture parameters.

From a practical standpoint, our method is suitable for large images (8k or more) from modern cameras. We achieve this by \rev{solving a smaller system of equations by collecting pixel-pairs with the lowest relative variances according to the noise model we consider.} Contrary to existing methods that utilize pixel pairs from consecutive exposures in the stack, \rev{we introduce a greedy algorithm that provides optimal pixel pairs to ensure that all exposures are well-estimated.} Our reduced linear system is a union of the highest weighted spanning trees induced on an \emph{exposure multigraph}, resulting in a balanced linear system.
% The efficient \emph{batched-MST} approach matches pixels from any image in the stack to the corresponding pixels in longer exposures since these are more reliable (lesser variance of noise). Finally, we utilize an iterative outlier removal procedure to make the solution robust to ghosting caused by camera or object motion.
\rev{Additionally, our proposed method is spatially balanced because we split the input image into tiles and collect pixels from all tiles. This improves the robustness of our estimator to ghosting caused by camera or object motion.}

\rev{Here is a brief overview of the paper that summarizes our contributions. We highlight that camera metadata may be unreliable and motivate the need to estimate corrective per-exposure ratios to obtain artifact-free \ac{HDR} reconstruction. We show that such exposure ratios can be estimated by solving a large linear system of equations (\secref{linear-system}), where each equation connects pixel intensities in the logarithmic domain. To deal with underexposed pixels in shorter exposures, we model heteroskedastic camera noise with inverse-variance weights (\secref{heteroskedastic}). Then, in \secref{reduced-system}, we show how to reduce the system of equations for faster equations without sacrificing the estimation quality. We finally validate our exposure estimation framework in \secref{results}, both on synthetic and real image captures. \footnote{\tcirev{Code for the method has been integrated into the software for noise-optimal HDR merging (\emph{HDRutils}) and can be found at \url{https://github.com/gfxdisp/HDRutils}.}}}

% \PH{Do we need list of contributions/high-level summary?}\RM{Having such a list of contribution is a "SIGGRAPH" style, but I do not see it often in other journals. You can check other papers in this journal.}

\section{Related work}
The problem of inaccurate capture parameters was identified in very early works in HDR merging \cite{debevec1997recovering,mitsunaga1999radiometric}. However, most of these focused on the challenging task of inverting the \ac{CRF} under the assumption of film or sensor \emph{reciprocity} \cite{tani1995photographic}. Mitsunaga and Nayar \cite{mitsunaga1999radiometric} used a polynomial model to jointly estimate the \ac{CRF} and exposure times. Then, Grossberg and Nayar \cite{grossberg2002brightness} demonstrated how to recover the \ac{BTF}, a function that describes how the brightness transfers from one image to another. They recovered the \ac{BTF} from image histograms and stipulated that it can be used to estimate exposure ratios if the \ac{CRF} is known. More recently, Rodríguez at al. have shown that the previously mentioned methods rely on incorrect assumptions \cite{Rodriguez2019} about the independence of color channels and linearity of exposures, resulting in estimation errors and hue shifts. We avoid those problems by directly operating on demosaiced RAW images. 

Based on \cite{grossberg2002brightness}, Cerman and Hlavac \cite{cerman2006exposure} assumed a linear \ac{CRF} by relying on RAW pixel values. They computed the brightness-transferred image histograms and used them to weight a system of equations in the linear pixel domain.

In contrast to their work, we solve a weighted linear system in the logarithmic domain to estimate exposure ratios. The weights in our system of equations are derived from a popular statistical camera noise model and ensure a noise-optimal solution. Our approach systematically handles heteroskedastic camera measurements and provides accurate estimates even when many pixels are under-exposed or affected by noise.

\subsection{Camera noise model}
A key contribution of our work is the use of a parametric noise model to weigh some pixel correspondences more than others. Our weights are based on the popular Poisson-normal statistical camera noise model~\cite{foi2008noise,aguerrebere2012study,konnik2014high} that has signal-dependent and static (or signal-independent) components. Many earlier works have used approximations of similar noise models for noise-optimal \ac{HDR} reconstructions \cite{hasinoff2010noise,granados2010optimal,hanji2020noise}. Although deep generative networks \cite{abdelhamed2019noiseflow,chang2020learning} model spatially-varying components of real camera noise better, the Poisson-normal statistical model and its normal approximation are better suited for our problem as they offer a convenient algebraic form under the assumption that noise is independent at each pixel.

\subsection{HDR datasets}
Merged HDR images of many multi-exposure datasets \cite{fairchild2007hdr,Karaduzovic-Hadziabdic2016,kalantari2017deep,hanji2021hdr4cv,Perez-Pellitero_2021_CVPR,hanji2022sihdr} can be improved with accurate exposure estimation. We observed the banding artifacts depicted in \figref{problem} in various images from these datasets. Recent HDR deep learning reconstruction methods for tasks like HDR deghosting \cite{kalantari2017deep,Pu2020robust,ye2021psfn,prabhakar2021self,chen2021attention} and inverse tone-mapping \cite{eilertson2017hdr,Marcel2020LDRHDR,wang2021deep} utilize these as well as other multi-exposure datasets for testing and evaluation. All these works are likely to produce better results when trained with accurate ground truth information due to our work on better exposure alignment.

\section{Methodology} \label{sec:method}
After introducing the camera model and some terminology, we describe how to estimate exposure ratios directly from the input image stack by solving a weighted linear system. We then derive noise-optimal weights for the system based on the widely-used Poisson-normal camera noise model. Finally, we discuss practical considerations for making the system
agnostic to sensitive noise parameters and improve robustness to ghosting caused by camera or object motion.

\subsection{Camera model}
To digitally represent an \ac{HDR} scene, we start by capturing $N$ images with varying exposure times or gains (ISO). Although earlier works included \ac{CRF} estimation in their image formation pipelines \cite{debevec1997recovering,mitsunaga1999radiometric,grossberg2002brightness}, we skip this step since modern cameras provide access to linear RAW pixel values. We model the captured RAW pixels as samples from independent random variables, which are linearly related to the scene radiance, and denote them as:
\begin{equation}
    Y_i(p), \quad i=1{\ldots}N, \quad p=1 \ldots M\,,
\end{equation}
where $i$ is the exposure index with $N$ total exposures, and $p$ is the pixel index with $M$ total pixels. Throughout this work, we use upper case for random variables and lower case for observed values. Before merging (averaging) the exposure stack, we need to convert them to relative radiance units by compensating for exposure time $t_i$, gain $g_i$, and aperture f-number $a_i$:
\begin{equation} \label{eq:exp-compens}
X_i(p) = \frac{Y_i(p)}{t_i\,g_i\,\pi\,\left(\frac{f}{2\,a_i}\right)^2} = \frac{Y_i(p)}{d_i}\,,
\end{equation}
where $f$ is the focal length. \tci{Typically, the scaling constant, $d_i = t_i \, g_i \, \pi \, \left( \frac{f}{2 \, a_i} \right)^2$, is computed directly from the camera \ac{EXIF} header. The problem is that $t_i$, $g_i$, and $a_i$ could be incorrect due to the limited accuracy of the mechanical shutter or the rounding of exposure values. Incorrect values of $d_i$ will lead to inaccurate HDR reconstructions} Thus, samples from $X_i(p)$, the exposure-compensated or absolute estimates in the captured images, may represent biased measurements of the true scene radiance. Merging images using the inaccurately reported parameters results in the banding artifacts depicted in \figref{problem} and \figref{gradient}. In all the images, the artifacts appear when a longer exposure image saturates.

\begin{figure}
    \centering
    \includegraphics[width=\columnwidth]{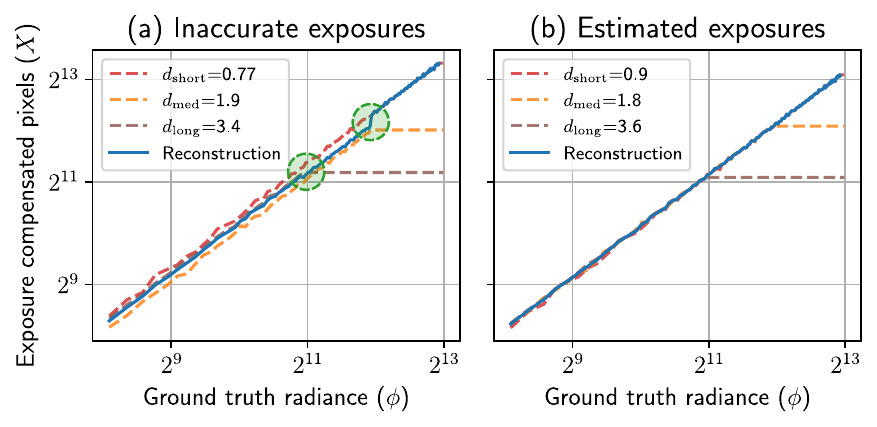}
    \caption{A linear gradient spanning $13$ stops is captured with exposure times $[0.5, 2, 4]$ (in seconds) by simulating the \emph{Sony ILCE-7R} at ISO $3200$. Each capture is quantized to $8$ bits for easy visualization. \rev{The left plot depicts exposure-compensated pixels that represent samples from $X(p)$ (according to \eqref{exp-compens}) with inaccurate exposures (red, yellow, and brown dashed lines). The reconstructed gradient (blue line) is jagged around $\phi=2^{11}$ and $\phi=2^{12}$ (green circles) due to misaligned exposures caused by biased exposure values.} We can reconstruct the smooth gradient by correctly aligning exposure ratios, as shown in the right plot.
    }
    \label{fig:problem-1d}
\end{figure}

\subsection{Banding due to inaccurate exposure}
\label{sec:banding}

To better understand the reason for banding, consider the one-dimensional linear gradient depicted in \figref{problem-1d} (left). The noisy measurements (dashed lines), obtained by simulating captures using calibrated noise parameters of the \emph{Sony ILCE-7R}, become misaligned when scaled with inaccurate exposure times. Merging such images results in a jagged reconstruction (solid blue line), causing banding in an otherwise smooth output. Notice that although the reconstruction deviates from the ground truth for almost all pixels, artifacts will be visible only at transition points when an image in the stack saturates \rev{as highlighted by the green circles.}

While \figref{problem-1d} demonstrates the problem with simple averaging, banding is further exaggerated when sophisticated algorithms based on physical noise models ~\cite{granados2010optimal,hasinoff2010noise,hanji2020noise} are used. This is because pixels of longer exposures are more reliable (due to smaller noise variance), but they saturate at lower physical values. Thus, these estimators weigh longer exposure pixels more, and there is a sharp change upon saturation of any image in the stack.

If we are able to estimate relative exposure ratios w.r.t one of the images correctly, we can align all exposures to produce the reconstruction in \figref{problem-1d} (right). Note that the exposure-aligned reconstruction may still not coincide with the ground truth since we do not have the correct baseline. However, accurately estimating relative ratios is sufficient to align the capture parameters and eliminate banding.

\subsection{Exposure estimation in real images}
\label{sec:linear-system}

To prevent banding and obtain physically accurate pixel values, we align the exposures of all images in the stack by estimating all scaling constants $d_i$.
% We start by treating the ratios as random variables and try to compute their expected values over all pixels.
% \begin{equation}
%     d_i = \E_p \left[ \frac{Y_i(p)}{X(p)} \right]\,,
% \end{equation}
% where $\E_p$ indicates expected value computed over all pixels.
% where $\E_p$ indicates that we will compute the expected value over all pixels.
% 
% \RM{This is unnecessary. Skip the equation above (Y/X) and directly introduce the one below (Y/Y).}
However, computing them directly from input pixels is impossible since the correct absolute measurements representing observations of $X(p)$ from \eqref{exp-compens} are unknown. We thus eliminate these unknowns by estimating the ratio of exposures between any two images in the stack instead. For a scene with constant illumination and no scene motion, this ratio should be the same for all pixels from a given pair of images. Its expected value is:
\begin{equation} \label{eq:expectation-ratio}
    d_{ij} = \E_p \left[ \frac{Y_i(p)}{Y_j(p)} \right]\,,
\end{equation}
where $i$ and $j$ index different images in the exposure stack, and $\E_p$ indicates that the expected value is computed over all pixels. To allow for fast computation using linear solvers, we operate on logarithmic values. Thus, let
\begin{equation} \label{eq:log-transform}
    e_{ij} = \log{d_{ij}} \quad \mathrm{and}\, L_i(p) = \log{Y_i(p)}\,.
\end{equation}
Although we cannot write a closed-form expression for the density function of $L_i(p)$ because of the $\log$ transformation, it is possible to approximate the expected value of any transformed random variable using its Taylor expansion. \rev{Our results, detailed in \eqref{appendix-ev} in the Appendix, are only applicable to normally distributed random variables. It is thus imperative to use only well-exposed pixels (we show how to do this in \secref{reduced-system}), since the Poisson photon noise component of such pixels is well-approximated by a normal distribution.} We can obtain an approximate expression for the expected value by applying the result for $\log{Y(p)}$:
\begin{equation} \label{eq:expectation-diff}
\begin{split}
    e_{ij} &= \log \E_p \left[ \frac{Y_i(p)}{Y_j(p)} \right] \approx \E_p \left[ \log{\frac{Y_i(p)}{Y_j(p)}} \right]\\
    &= \E_p \left[ L_i(p)-L_j(p) \right]\,.
\end{split}
\end{equation}
The equality approximately holds for the operating range of pixel values, as we will detail in \secref{heteroskedastic}, \eqref{log-moments}. This allows us to compute the expected value over all pixels by setting up and solving a linear system. For the most reliable estimate, we should utilize information from all the available exposures (all possible values of $i$ and $j$). Unlike previous work~\cite{cerman2006exposure}, we thus consider not only ratios between neighboring exposures but between all pairs. This results in a large yet sparse linear system:
\begin{gather} \label{eq:exp-sparse-lin}
    \sqrt{\bm{W}} {\tiny \begin{bmatrix}
        1 & -1 & 0 & \cdots & 0 \\
        1 & -1 & 0 & \cdots & 0 \\
        \vdots & \vdots & \vdots & \ddots & \vdots \\
        1 & 0 & -1 & \cdots & 0 \\
        \vdots & \vdots & \vdots & \ddots & \vdots \\
        0 & 0 & 0 & \cdots & -1 \\
    \end{bmatrix}
    \begin{bmatrix}
        e_1 \\ e_2 \\ \vdots \\ e_N
    \end{bmatrix}} =
    \sqrt{\bm{W}} {\tiny \begin{bmatrix}
    L_1(1)-L_2(1) \\
    L_1(2)-L_2(2) \\
    \vdots \\
    L_1(1)-L_3(1) \\
    \vdots \\
    L_{N-1}(M)-L_N(M) \\
    \end{bmatrix}}\,,
\end{gather}
or more compactly,
\begin{equation} \label{eq:wls}
    \sqrt{\bm{W}} \bm{O} \bm{e} = \sqrt{\bm{W}} \bm{m}\,,
\end{equation}
where $\bm{W}$ is a diagonal weight matrix denoting the relative importance of each row of the system.
% The required exposure ratios are extracted from the closed-form solution:
% \begin{equation} \label{eq:closed-form}
%     \bm{\tilde{e}}_\mathrm{WLS} = (\bm{O}\transp \bm{W} \bm{O})^{-1} \bm{O}\transp \bm{W} \bm{m} \,.
% \end{equation}
%
In practice, we found that the weighted system does not always provide a good solution for shorter exposures. Since a rough estimate of exposure values $\bm{e}_0$ is available in the image metadata, we introduce a Tikhonov penalty with weight $\lambda$ and solve to get:
% and instead minimize the following expression:
% \RM{The system below is still linear. Why do you need an iterative solver? This is identical to what Gabriel used in the SI-HDR paper}:
\begin{equation} \label{eq:lsqr}
\begin{split}
    \bm{\hat{e}}_\mathrm{WLS} &= \argmin_{\bm{e}}\, \left \lVert \sqrt{\bm{W}} (\bm{O}\bm{e} - \bm{m}) \right \rVert^2_2 + \lambda \lVert \bm{e} - \bm{e}_0 \rVert^2_2\\
    &= (\bm{O}\transp \bm{W} \bm{O} + \lambda \bm{I})^{-1} (\bm{O}\transp \bm{W} \bm{m} + \lambda \bm{e}_0)\,.
\end{split}
\end{equation}
% After regularization, the solver~\cite{fong2011lsmr} converges rapidly and requires fewer than 10 iterations even for large images.

\begin{figure}
    \centering
    \includegraphics[width=.49\columnwidth]{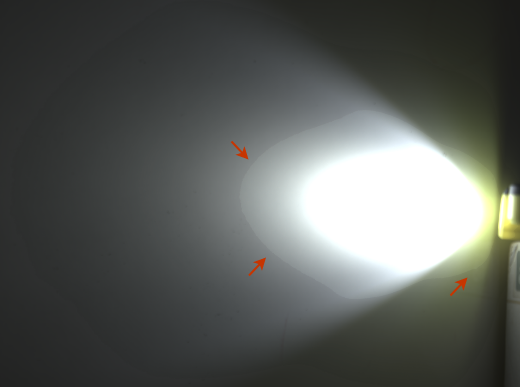}\hfill
    \includegraphics[width=.49\columnwidth]{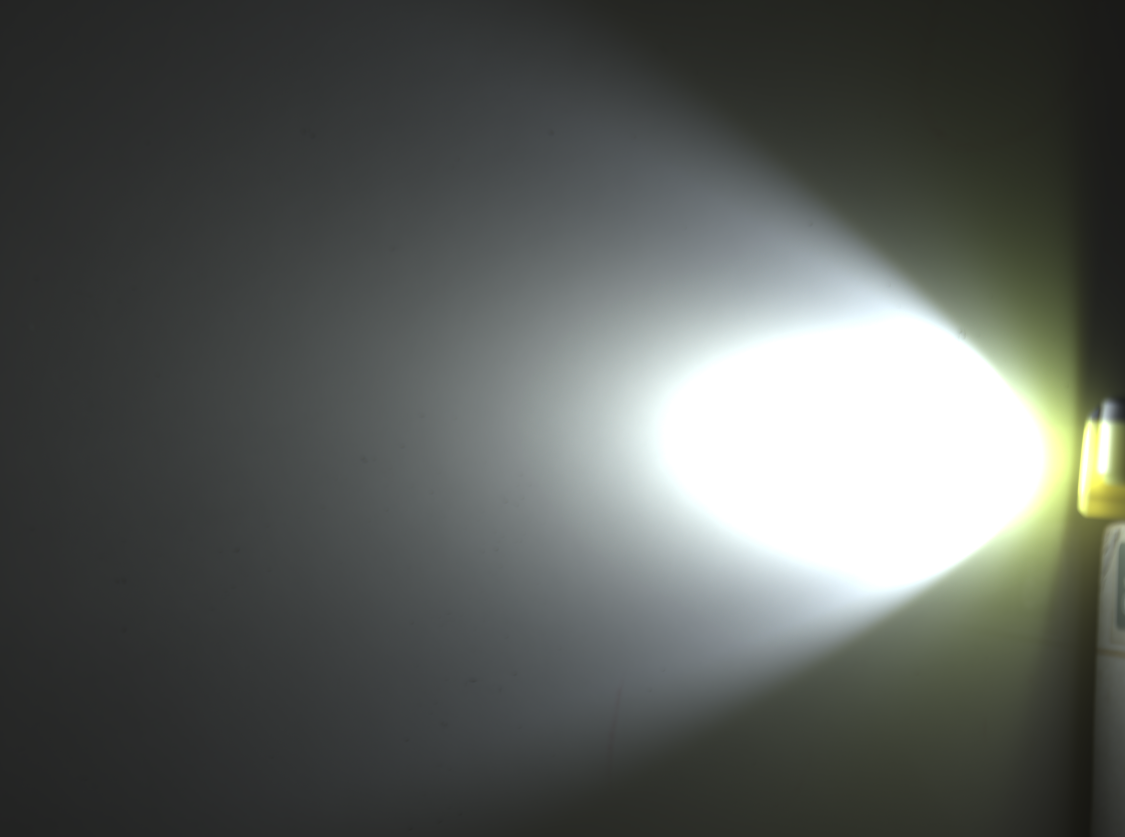}\\[-2ex]
    \subfloat[Parameters from \ac{EXIF}]{\includegraphics[width=.49\columnwidth]{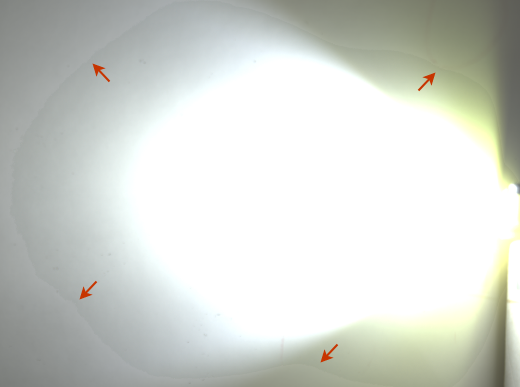}}\hfill
    \subfloat[Estimated parameters]{\includegraphics[width=.49\columnwidth]{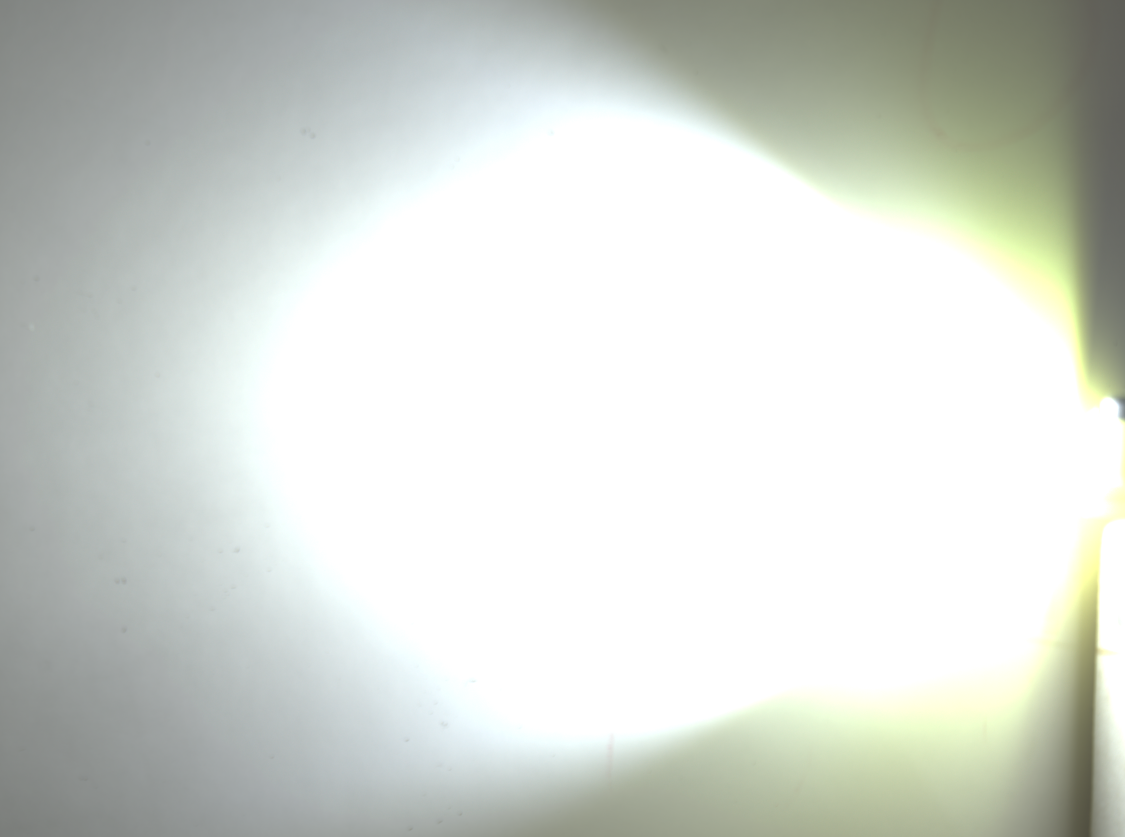}}
    \caption{
    % For visualisation, we show two exposures (top and bottom rows) of \ac{HDR} images reconstructed using different sets of exposure ratios.
    The left column shows exposures (gamma-encoded for visualization, $\gamma=2.2$) of the HDR image reconstructed with \ac{EXIF} parameters, while the right column shows exposures (encoded with $\gamma=2.2$) of the HDR image reconstructed by solving the linear system represented by \eqref{exp-sparse-lin} with $\bm{W} = \bm{I}$. Banding artifacts are visible in dark (top row) and bright (bottom row) regions of the reconstruction using EXIF metadata (a). Aligning exposures according to \eqref{lsqr} fixes the problem (b).}
    \label{fig:gradient}
\end{figure}

% Earlier methods \cite{grossberg2002brightness,cerman2006exposure} recommend constructing a similar system with weights given by scaled histogram values. On the contrary, we derive weights with the help of a statistical noise model in \secref{heteroskedastic}. Finally, using the $l2$ loss, the vector of $log$ exposure ratios $\bm{e}$ is simply the least squares solution of the system. As shown in \figref{gradient}, for simple scenes, we can then use $\exp(\bm{e})$ to estimate new relative exposure times and use those in place of \ac{EXIF} data to merge the \ac{HDR} image stack.

% \subsubsection{Homoskedastic pixels}

% \RM{I would remove this section. There is no point in modeling unrealistic camera. Jump directly to the explanation of the actual noise model to make the paper shorter and more straightforward. }
% \PH{I think it is good to retain it since it motivates the next section better. Specially since there are examples which can be solved by this simple approach.}
% If the noise levels of most pixels are much smaller than the signal, it is possible to get a good estimate of exposures by assuming that all the rows of \eqref{exp-sparse-lin} are equally reliable or homoskedastic (affected by the noise with constant variance). This corresponds to setting the weights to an identity matrix $\bm{W}=\mathbb{I}$. Then, from the ordinary least-squares solution $\bm{\hat{e}_\mathrm{OLS}}$, we recover exposure ratios by exponentiation $\bm{\hat{d}_\mathrm{OLS}} = \exp{\bm{\hat{e}_\mathrm{OLS}}}$.

To demonstrate the effectiveness of exposure alignment, we captured an exposure stack of a simple \ac{HDR} scene consisting of a bright light shining at an angle, which produces an inverse-square fall-off in intensity. Since most pixels are well-exposed, as shown in \figref{gradient}, we could assume that they are equally reliable and set $\bm{W} = \bm{I}$. This works well for the carefully controlled scene and eliminates banding artifacts that would have appeared when merging with \ac{EXIF} parameters. However, the constant noise assumption breaks down for real-world HDR stacks since the noise in camera pixels is heteroskedastic~\cite{foi2009heteroskedastic} and thus, different pixels provide different amounts of information.

% For the simple gradient depicted in \figref{gradient}, such a system correctly estimates the true exposure ratios and eliminates banding artifacts that would have appeared when merging with \ac{EXIF} values. Notice the banding artifacts visible in images in the left column due to misalignment of exposures when using values reported in \ac{EXIF} data. After estimating the true exposure ratios by solving \eqref{closed-form} with identity weights, we obtain the clean images in the right column.

\subsection{Heteroskedastic pixels} \label{sec:heteroskedastic}
To determine noise-optimal weights for images of real cameras, we need to derive an expression for the variance of each row of \eqref{exp-sparse-lin}. Then, we populate $\bm{W}$ with inverse-variance weights (i.e., each weight is given by the reciprocal of variance) to obtain the weighted least-square estimate $\bm{\hat{e}_\mathrm{WLS}}$. This is equivalent to \ac{MLE} under the assumption that the system of equations models additive, normally distributed noise, where the variance is different for different rows. Since each entry of the output vector $\bm{m}$ is a difference of two random variables, the inverse-variance weight for each row of the system is given by:
\begin{equation} \label{eq:w-inv-var}
    w_{k,k} = \frac{1}{\V[L_i(p) - L_j(p)]} = \frac{1}{\V[L_i(p)] + \V[L_j(p)]} \,.
\end{equation}
Here, $k$ indexes each row of \eqref{exp-sparse-lin} since $\bm{W}$ is diagonal. For instance, the second row corresponds to $k = 2, i = 1, j = 2$.

In order to get an expression for the denominator in \eqref{w-inv-var}, we refer to detailed studies of the noise characteristics of cameras \cite{aguerrebere2012study,konnik2014high}, which indicate that real camera noise follows a compound Poisson-normal distribution. For the working range of pixels in most images, this can be approximated by zero-mean additive noise that follows a normal distribution. Temporarily dropping the exposure index for brevity of notation, the variance at each pixel thus consists of a signal-dependent component as well as a static component and is equal to,
\begin{equation}
    \sigma_Y^2(p) = \V[Y(p)] = \alpha\,\E[Y(p)]+\beta\,,
    \label{eq:noise-model}
\end{equation}
where $\alpha$ and $\beta$ are camera-specific noise parameters, which also depend on the sensor's gain.

Since $L(p)$ is a random variable obtained by applying the $\log$ transformation to $Y(p)$, its density function does not have an exact expression. This is because the domain of $Y(p)$ includes negative values for which the $\log$ function is undefined. \rev{As we will show in \secref{reduced-system}, we operate on a small subset of available pixels, selected for the lowest relative variances, and whose intensities tend to be much greater than $0$.} We show, in \eqref{appendix-var} in the Appendix, how to approximate the variance when any random variable is transformed by an invertible function. Here, we apply the result for $Y(p)$ when it is transformed by the $\log$ function,
% \ac{MLE} tells us that the expected value represented by \eqref{wls} is equivalent to a weighted a average where weights are given by the inverse variance sum,
% \begin{equation} \label{eq:w-inv-var}
%     w_{k,k} = \frac{1}{\V[L_i(p) - L_j(p)]} = \frac{1}{\V[L_i(p)] + \V[L_j(p)]} \,.
% \end{equation}
% Here, $k$ indexes each row of \eqref{exp-sparse-lin} since $\bm{W}$ is diagonal; $i$ and $j$ index images in the stack. For instance, the second row corresponds to $k = 6, i = 0, j = 1$. To combine \eqref{noise-model} and \eqref{w-inv-var}, the final step is to express $\V[L(p)]$ in terms of $\sigma^2_Y(p)$. From \eqref{appendix-ev-var}, we have:
\begin{equation} \label{eq:log-moments}
\begin{split}
    & \E[L(p)] \approx \log{\mu_Y(p)}\,,\\
    & \V[L(p)] \approx \frac{\sigma^2_Y(p)}{\mu^2_Y(p)}\,,
\end{split}
\end{equation}
where $\sigma^2_Y(p)$ is the variance of the noise from \eqref{noise-model} and $\mu^2_Y(p)$ is the expected pixel value, which we approximate by $\E[Y(p)] = \mu_Y(p) \approx y(p)$. \rev{We reiterate that we are able to use this result because we work with well-exposed pixels, that are approximately normal.}
%For a given exposure and pixel location, there exists a single observation which serves as the mean or expected value i.e. $\mu_Y(p) = y(p)$, while its variance is given by \eqref{noise-model}. 
After substituting the computed variance in \eqref{w-inv-var}, the diagonal weights become:
\begin{equation} \label{eq:weights}
    w_{k,k} = \left(\frac{\alpha y_i(p) + \beta}{y_i^2(p)} + \frac{\alpha y_j(p) + \beta}{y_j^2(p)}\right)^{-1}\,.
\end{equation}

\subsection{Camera noise calibration} \label{subsec:no-camera}

While the inverse-variance weights in \eqref{weights} help us compute noise-optimal estimates of the exposure ratios, a fundamental limitation is their dependence on calibrated noise parameters $\alpha$ and $\beta$. In the considered noise model \cite{foi2008noise}, these are camera and gain specific and may not be available at the time of \ac{HDR} merging. Moreover, the quality of \ac{HDR} reconstructions is highly sensitive to accurate noise parameters~\cite{aguerrebere2012study}, motivating the need for methods that do not rely on them.

If noise parameters are unavailable or inaccurate, it is still possible to solve the weighted linear system by assuming that the static noise parameter $\beta$ is $0$ as we work with well-exposed pixels (see \secref{reduced-system}). Since the entries of $\bm{W}$ determine the relative importance of the different rows of ~\eqref{exp-sparse-lin}, we can eliminate the common signal-dependent constant, $\alpha$ as well. The new weights which no longer depend on calibration-sensitive parameters are:
\begin{equation} \label{eq:weights-prime}
    w'_{k,k} = \left( \frac{1}{y_i(p)} + \frac{1}{y_j(p)} \right)^{-1}\,.
\end{equation}

In practice, we can get away with using camera-independent $\bm{W'}$ instead of camera-specific $\bm{W}$ from \eqref{weights}.

\subsection{Reducing the linear system} \label{sec:reduced-system}
The linear system given by \eqref{exp-sparse-lin} contains up to $M\binom{N}{2}$ equations corresponding to all pixels ($M$) and all pairs of exposures in the stack ($N$). Solving such a large system may be impossible for large images or deep exposure stacks due to computation and memory limitations. However, this system is strongly overdetermined as only $N-1$ exposure ratios need to be estimated. Therefore, we only need a small percentage of equations to solve \eqref{lsqr}. \rev{Another reason for reducing the system is to eliminate underexposed pixels pairs because they do not satisfy the assumptions made in the derivations of previous sections. For example, dark pixels may be negative or poorly approximated by a normal distribution.}

A logical reduction strategy is to select pixels pairs with the highest weights since they are least affected by noise. \rev{However, such a strategy is heavily biased towards longer exposures because pixel pairs that include these images will have the highest weights. The shorter exposures will then be poorly represented and, thus, poorly estimated. Another issue is related to the spatial location of bright objects in the scene, such as the Sun or other light sources. If we select a small fraction of pixels (say 5\% of the pixels per image), all of them are likely to be concentrated in one portion of the image, corresponding to these bright objects. If those objects happen to be in motion, the exposure estimation will fail.}

\rev{We propose two orthogonal design choices to balance the system of equations and handle both these biases in estimated exposure ratios.}

\subsubsection{Spatial balance: Tiling}
\label{sec:tiled}
\rev{To ensure that the linear system contains samples from all parts of a scene, we split the input image stack into $t \times t$ pixel tiles ($16 \times 16$ in our experiments). We can then select a fixed number of pixel pairs from each tile and pixels from a few bright objects will not dominate the system of equations.}

\rev{The tiled processing provides a convenient way to vectorize the construction of the reduced linear system. Several tiles can be processed in parallel for faster execution with multi-core or multi-threaded systems.}

\rev{Our noise-based solution helps provide a robust estimation even though some tiles may contain noisy pixels corresponding to dimly-lit parts of the scene. Variance-optimal balancing of exposures is crucial for such tiles, as we will show in the next section.}

\subsubsection{Exposures balance: Spanning trees}
\label{sec:mst}

\begin{figure*}
    \centering
    \includegraphics[width=\textwidth]{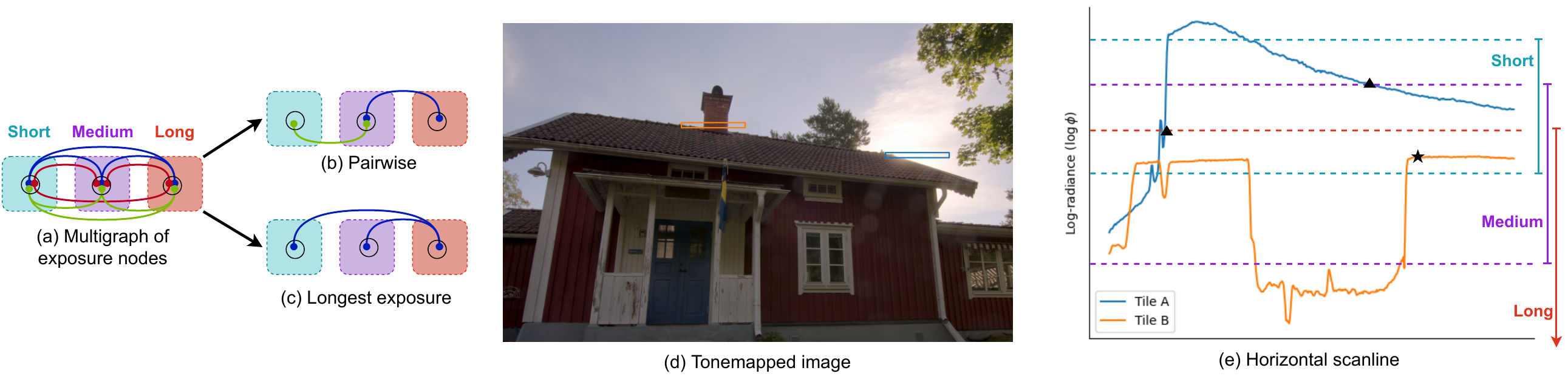}
    \caption{\rev{Consider the exposure multigraph in (a) where different colored vertices and edges represent different pixels. The \ac{MST} may connect pairwise exposures (b) or connect the longest exposure to all others (c). \algref{greedy-mst} will provide different solutions for different inputs. For example, the \ac{MST} for the blue horizontal tile near the Sun in (d) is the pairwise-connected spanning tree connected at pixels marked by the two triangles in plot (e). However, for the orange tile, the \ac{MST} is given by the longest-exposure spanning tree at the pixel marked by the star in plot (e).}}
    % \RM{It may not be clear that the dots in (a) represent three pixels at different spatial locations. Do colours correspond to those in (e)). If yes, state so. Perhaps instead of showing multiple MSTs in (b) and (c), show just one MST based on your example in (e).}}
    \label{fig:exp-graph}
\end{figure*}

\rev{The next objective is guaranteeing that all $N-1$ exposure ratios are correctly estimated. Within each spatial tile, we need to include pixels from all exposure pairs, including noisy short exposures despite their relatively smaller weights.}

\rev{Consider the \emph{exposure graph}: an undirected weighted multigraph (a graph that contains more than one edge between two vertices as shown in \figref{exp-graph}a), whose vertices represent the $N$ exposures and edges link pairs of co-located pixels from two exposures. Each edge corresponds to one row of \eqref{exp-sparse-lin}, with its weight given by \eqref{weights} or \eqref{weights-prime}. The different colored vertices and edges in \figref{exp-graph}a represent different pixel locations. Reducing the linear system from \eqref{exp-sparse-lin} is then equivalent to removing edges from this dense multigraph.}

\rev{The optimal subset contains $(N-1)k$ edges with the largest weights that connect all exposures in a balanced manner. Such a subset can be found by computing the $k$ highest-weighted spanning trees of the multigraph, where the weight of a spanning tree is the sum of the weights of its edges. The previous work directly linked pixels from neighboring exposures~\cite{cerman2006exposure}, resulting in pairwise connectivity as shown in \figref{exp-graph}b. However, this solution is sub-optimal for some inputs (such as the orange tile in \figref{exp-graph}d) because using edges linked to the longest exposures, as shown in \figref{exp-graph}c), results in higher weights and, therefore better estimates.}

\rev{An optimal solution would sequentially extract $k$ \ac{MST}s using \emph{Kruskal's} or \emph{Prim's} algorithm. Better algorithms~\cite{katoh1981algorithm,eppstein1990finding} extract the $k$ highest spanning trees more efficiently. However, explicitly creating a large multigraph and computing \ac{MST}s is both memory and computationally expensive. Below we show that the optimal solution can be found by a simpler greedy algorithm.}

% \RM{The way I understand it: the position of the brightest pixels in a given exposure is also the position of the brightest pixel in a longer exposure as long as the pixel in the longer exposure is not saturated (it would be good to show a plot demonstrating that). Therefore, an MST can be constructed by selecting the brightest pixel in the shortest exposure and connecting it to the corresponding pixels in the longer exposures up to the exposure $k$, in which this pixel location is saturated. Then, we repeat procedure starting with the exposure $k-1$. Once we connect all exposures, we construct an MST without explicitly building a graph. It could happen that some exposures are ignored in a given tile because all pixels are saturated. In that case, the connection between exposures should be made in other tiles.}

\paragraph*{Greedy MST solution}

\begin{algorithm}
\SetKwInOut{Input}{input}\SetKwInOut{Output}{output}
\Input{$y_{1}, \ldots, y_{N}$ (Images sorted by exposure time)}
\Output{$\textrm{MST}$ (Maximimum spanning Tree)}
\DontPrintSemicolon
\caption{Greedy MST algorithm} 
\label{algo:greedy-mst}
\SetKwFunction{FMST}{greedyMST}
\SetKwFunction{FWeights}{maxWeight}
\SetKwFunction{FValid}{isValid}
\SetKwFunction{FAppend}{addEdge}
\SetKwProg{Fn}{Function}{}{}
\BlankLine
\Fn{\FMST{$y[\;]$}}{
    $N \gets length(y)$ \;
    $\textrm{MST} \gets [\;]$ \;
    \For{$i \gets 1$ to $N-1$} {
        \tcc*[l]{Iterate over all exposures starting from the shortest}
        $mask \gets$ \FValid{$y_i$} and \FValid{$y_{i+1}$}\;
        $p_* \gets$ \FWeights{$y_i[mask], y_{i+1}[mask]$} \tcp*{Location of the highest weighted edge connecting images $y_i$ and $y_{i+1}$}
        \For{$j \gets N$ to $i+1$} {
            \If{\FValid{$y_j(p_*)$}} {
                $\textrm{MST}.$\FAppend{$i \edge j$} \;
                break \;
            }
        }
    }
    \KwRet {$\textrm{MST}$}
}
\end{algorithm}
% \RM{I would not be able to reproduce the algorithm from your description below. You do not state whether $i+1$ is longer or shorter exposure. "a linear search" is ambiguous. An algorithm must be explained with mathematical precision. Please create a figure with pseudocode and try rewriting the text below. Once I see the pseusocode, I will be able to improve the text.}
\rev{We start by extracting all valid pixels --- those that are unsaturated and sufficiently above the noise floor. We iterate through all exposures from the shortest to the longest.
% in ascending order \RM{ascending is unclear. Say from the shortest to the longest exposure if you mean that.}.
For each exposure $i$, we identify $p_*$, the pixel location that contains the highest-weighted edge between exposures $i$ and $i + 1$. This is typically the brightest pixel that is not saturated in exposure $i + 1$. Then, we find the longest exposure $j>i$ in which pixel $p_*$ is not saturated. We add an edge between $i$ and $j$. By selecting the longest exposure, we ensure that the weight of the edge is maximized.}

% \rev{If some longer exposure, $f > i + 1$ is valid at location $p_*$, the weight of edge $y_i(p_*) \edge y_f(p_*)$ will be larger than that of edge $y_i(p_*) \edge y_{i+1}(p_*)$. In fact, we can greedily search for the longest valid exposure by starting with the longest exposure, and iterating through the stack in descending order.
\rev{This procedure, summarized in \algref{greedy-mst}, can be repeated $k$ times to extract the $k$ highest-weighted spanning trees without explicitly constructing a graph. If all longer exposures $y_{N}(p_*), y_{N-1}(p_*), \ldots y_{i+2}(p_*)$ are invalid due to saturation, the solution reduces to pairwise connectivity depicted in \figref{exp-graph}b.}

\rev{Note that while a pair of pixels forming an edge must have the same position, each edge in the \ac{MST} can come from a different pixel position. Consider the 1-dimensional blue tile in the tonemapped image in \figref{exp-graph}d and its horizontal scanline in \figref{exp-graph}e. The \ac{MST} is given by two edges: an edge between the medium and long exposures and an edge between the medium and short exposures (marked by triangles in the plot). Note that these are the brightest unsaturated pixels in the long and medium exposures. The \ac{MST} for the blue tile is thus the pairwise connected spanning tree. The limited dynamic range of the orange tile means that a single pixel, marked by the star in \figref{exp-graph}e, aligns all three exposures for the orange tile, resulting in the spanning tree in which the longest exposure is linked to all other exposures.}

\subsection{Handling outliers and pixel misalignment} \label{sec:outlier}
\begin{figure}
    \centering
    \includegraphics[width=.49\columnwidth]{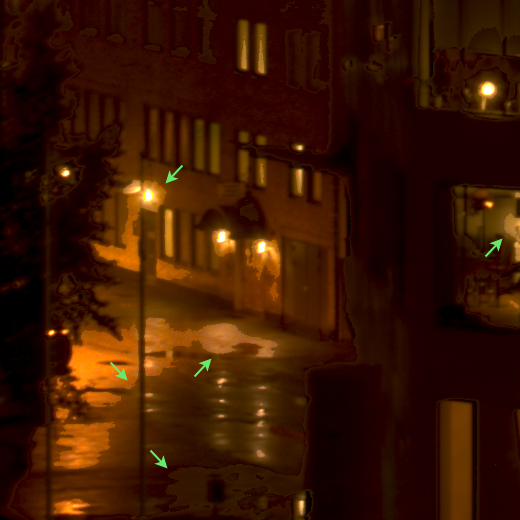} \hfill
    \includegraphics[width=.49\columnwidth]{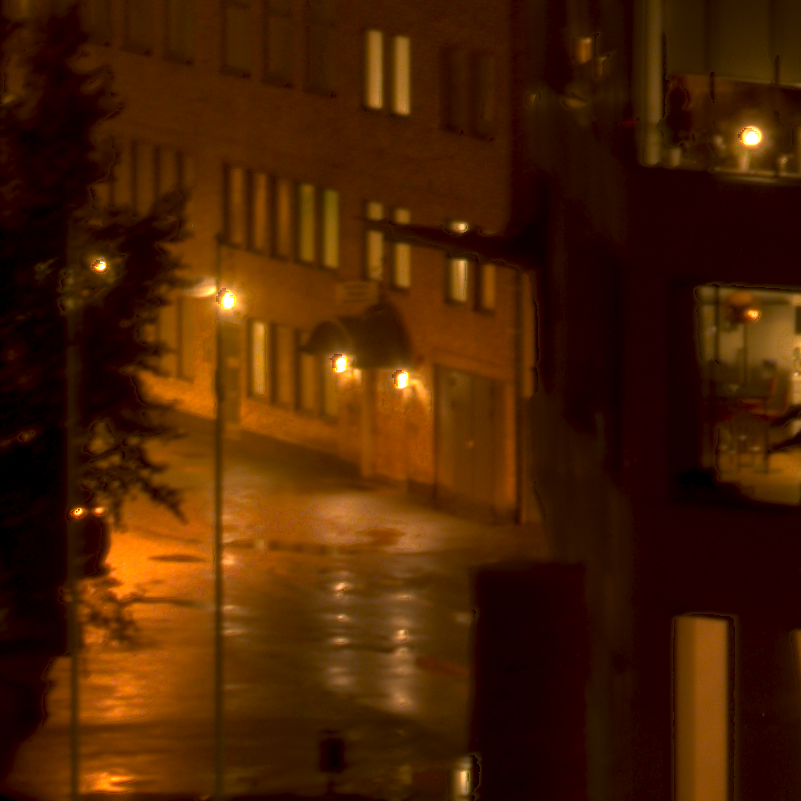} \\
    \subfloat[Regular $\bm{\hat{e}_\mathrm{WLS}}$ \label{fig:robust-base}]{\includegraphics[width=.49\columnwidth]{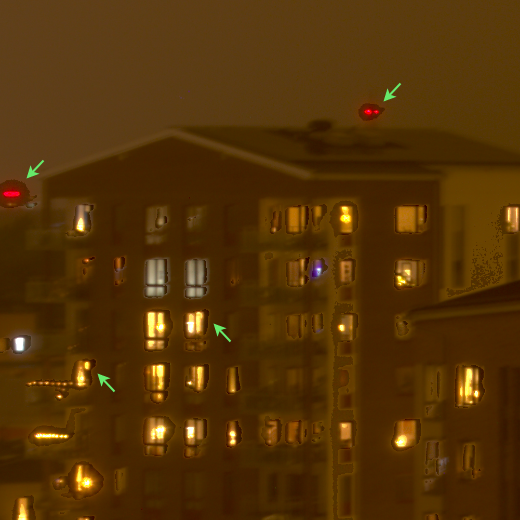}} \hfill
    \subfloat[$\bm{\hat{e}_\mathrm{WLS}}$ with outlier removal \label{fig:robust-outlier}]{\includegraphics[width=.49\columnwidth]{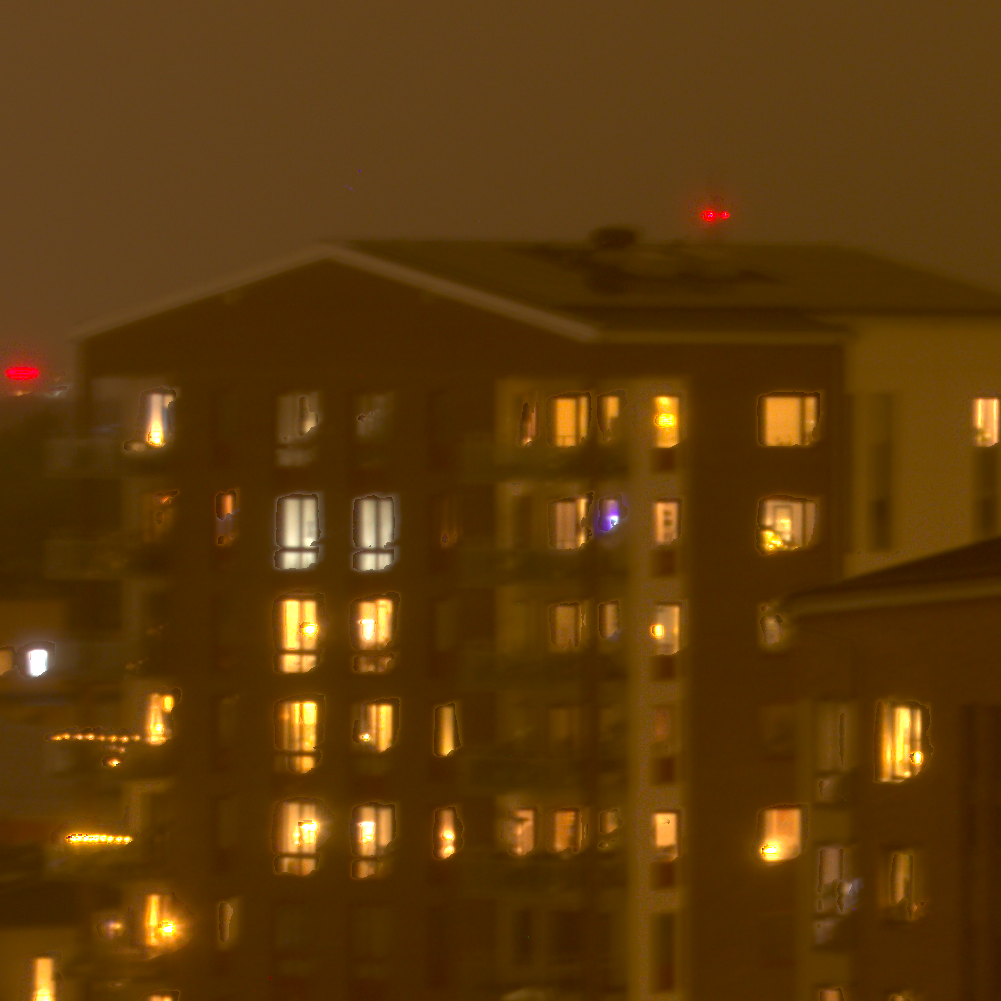}}
    \caption{For an exposure stack containing some spatially misaligned pixels, solving the weighted linear system \eqref{wls} results in inaccurate estimates. In this scene, camera motion causes pixels of the building to be misaligned due to camera motion, producing artifacts near the windows in \figref{robust-base}. However, pixels from other areas, such as the sky, can still be utilized for exposure estimation. \figref{robust-outlier} demonstrates the advantage of using the outlier-removal procedure described in \secref{outlier} to improve robustness.}
    \label{fig:robustness}
\end{figure}

The weighted least-squares solution \eqref{lsqr} using rows of the reduced system will provide an accurate estimate only if there is no movement in the scene and all pixels are well aligned. If there is motion across exposures, for example, due to camera shake or object motion, those pairs of pixels will affect our estimates of exposure ratios, as shown in \figref{robustness}. Notice the artifacts surrounding bright regions such as the windows and red lights in \figref{robust-base} due to camera motion. Here, we estimated the exposures using the original stack and then registered the images with homography alignment~\cite{Tomaszewska2007homography} before merging. Thus, any visible artifacts are due to incorrect exposure alignment and not due to ghosting.

\rev{The spatial tiling described in \secref{tiled} ensures that motion in a few bright objects does not adversely affect the estimated exposures, i.e., our procedure is robust to localized object motion. In case of widespread pixel misalignment (e.g., due to camera motion), we can still provide reasonable exposure estimates if the scene contains uniform image patches. We identify \emph{usable} tiles by collecting $k$ \ac{MST}s within a tile and solving the smaller per-tile system of equations separately. If the predictions of a tile deviate too much from the \ac{EXIF} values, we treat the tile as an outlier and do not include the corresponding \ac{MST}s in the final system.}

\rev{The added overhead of solving per-tile linear systems is negligible because each system contains only $k$ equations ($k = 50$ in our experiments). This outlier removal is much faster than an iterative algorithm such as the one described in \cite{cerman2006exposure}.}

% To make the system robust to such outliers, we employ an iterative algorithm similar to the one described in \cite{cerman2006exposure}. We first estimate exposures $\bm{\hat{e}_\mathrm{WLS}'}$ using the rows that remain after batched-MST reduction described in \secref{reduced-system} by solving for \eqref{lsqr}. Next we compute the weighted residual vector for the system,
% \begin{equation}
%     \bm{r} = \sqrt{\bm{W}} \bm{O} \bm{\hat{e}_\mathrm{WLS}'} - \sqrt{\bm{W}} \bm{m}\,.
% \end{equation}
% Finally, we further reduce the linear system by eliminating all rows whose weighted residuals lie more than $n$ standard deviations away from the mean of $\bm{r}$ ($n=3$ in our results). This marks the beginning of the next iteration, which improves the estimated $\bm{\hat{e}_\mathrm{WLS}'}$ by solving the reduced system, again according to \eqref{lsqr}. The algorithm converges when the change in the mean of weighted residuals is smaller than an $\epsilon$. The described procedure is robust to camera motion, as shown in \figref{robustness}, provided there are uniform patches in some regions of the scene. Similarly, the procedure will eliminate pixels corresponding to any moving objects from the estimation.

% It is worth noting that the above procedure works even if the number of outliers (misaligned pixels) is greater than the number of inliers. This is because outlier removal retains both under- and over-estimated exposures, which cancel out. Other popular robust estimators, such as \ac{RANSAC}, cannot deal with a large percentage of outliers.

\section{Results and applications} \label{sec:results}
The primary application of our method is merging exposure stacks with varying exposure times or gains. We first validate our results on synthetically generated stacks, for which we know the ground truth, and then show qualitative comparisons on real captures. Finally, in \secref{display-mtf}, we show how our method can improve the estimate of the MTF of a display. All our results use $\lambda=10$ for the Tikhonov penalty term described in \eqref{lsqr}. \tci{For merging exposures stacks after exposure estimation, we used a noise-aware HDR estimator~\cite{hanji2020noise}. However, our results hold for other simpler methods too.}

\subsection{Validation on synthetic dataset} \label{sec:res-synthetic}

\begin{figure}
    \centering
    \includegraphics[width=\columnwidth]{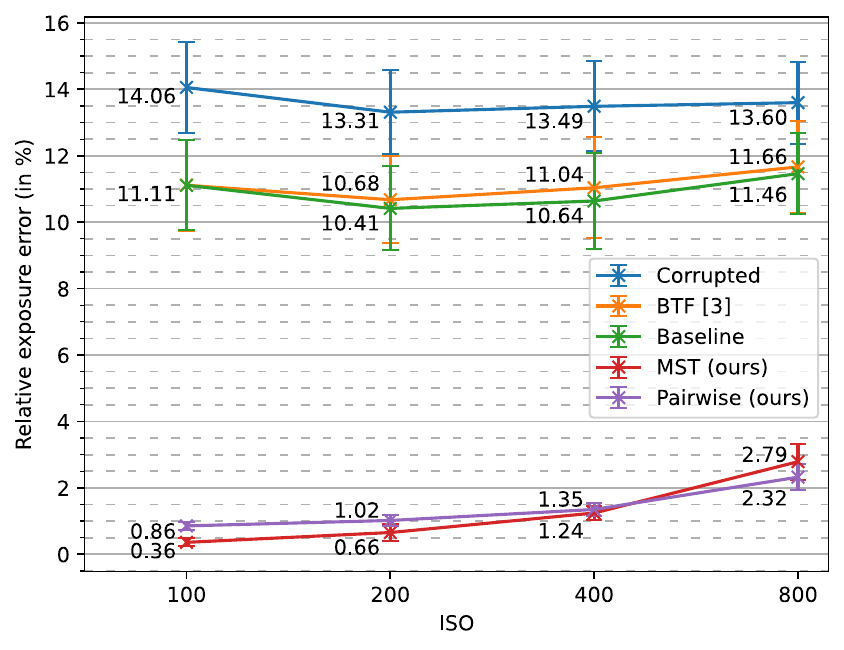}
    \caption{Accuracies of exposure estimation methods with increasing ISO computed by simulating captures on an existing HDR dataset~\cite{fairchild2007hdr}. Error bars indicate $95\%$ confidence intervals computed across all images. The blue line represents corrupted exposures according to \eqref{res-error}, \tci{while the baseline, in green, was obtained by directly estimating exposures by solving a system based on \eqref{expectation-ratio}.}
    }
    \label{fig:results-synthetic}
\end{figure}

First, we rely on synthetic exposure stacks to compare the accuracy of different methods for exposure estimation. We simulated \ac{HDR} exposure stacks using the noise parameters of the \emph{Canon PowerShot S100} with exposure times \rev{[\nicefrac{1}{64}, \nicefrac{1}{8}, 1, 8] (in seconds). All images were quantized to 14 bits to match the bit-depth of the camera}, and the experiment was repeated for ISO settings between 100 and 800. \tci{We simulated noise according to \eqref{noise-model}, with noise parameters listed in \tableref{noise-params}}.\footnote{\tcirev{Code for noise simulation is available at \url{https://github.com/gfxdisp/HDRutils/tree/main/HDRutils/noise_modeling}.}}

\begin{table}[t]
\centering
\caption{\tci{Noise parameters measured from the \emph{Canon PowerShot S100} at various ISOs.} \tcirev{These parameters have been calculated after normalizing the camera sensor values to a range of 0 to 1.}}
\label{tab:noise-params}
\begin{tabular}{@{}ccccccc@{}}
\toprule
ISO & $\alpha_\textrm{R}$ & $\alpha_\textrm{G}$ & $\alpha_\textrm{B}$ & $\beta_\textrm{R}$ & $\beta_\textrm{G}$ & $\beta_\textrm{B}$ \\ \cmidrule(r){1-1} \cmidrule{2-4} \cmidrule(l){5-7}
100 & 2.46e-5            & 1.67e-5            & 7.41e-5            & 3.58e-8            & 2.13e-8            & 1.28e-7            \\
200 & 4.57e-5            & 3.02e-5            & 1.32e-4            & 9.89e-8            & 6.07e-8            & 2.66e-7            \\
400 & 9.12e-5            & 5.95e-5            & 2.59e-4            & 2.21e-7            & 1.72e-7            & 5.61e-7            \\
800 & 1.85e-4            & 1.19e-4            & 5.26e-4            & 4.94e-7            & 4.28e-7            & 1.14e-6            \\ \bottomrule
\end{tabular}
\end{table}

Source \ac{HDR} images were taken from the Fairchild photographic survey~\cite{fairchild2007hdr} containing $105$ scenes at a resolution of $4312 \times 2868$. \rev{All simulated captures are linearly related to the HDR image, i.e., we do not apply a \ac{CRF} since it is likely to introduce artifacts.}

Before merging the images, we corrupted the exposure times by introducing a small amount of normally distributed noise:
\begin{equation} \label{eq:res-error}
    e'_i = e_i + \eta_i \quad \mathrm{where}\,\,\eta_i \sim \mathcal{N}(0, 0.15\,e_i)\,.
\end{equation}
\rev{The relative standard deviation factor $0.15$ was selected to match real camera \ac{EXIF} errors plotted in \figref{error-ratios}.}
% For both tested datasets, the magnitude of sample standard deviation of relative errors was close to $15\%$.

\rev{\figref{results-synthetic} plots the relative \ac{RMSE} (in percent) of the exposure ratios, for different ISO levels. Note that we can compute such relative errors only for synthetic datasets.  \tci{We include results of a simple baseline by solving a linear system that realises \eqref{expectation-ratio} without relying on a noise model. Note that the baseline includes other components described in \secref{method}, such as spatial balance via tiling and exposure balance via \ac{MST}s.} The blue line shows the error of noisy exposures according to \eqref{res-error}. The synthetic exposure stacks represent a challenging estimation problem with exposures three stops apart. \tci{As a result, the simple baseline (green) and the histogram-based \ac{BTF} method~\cite{cerman2006exposure} (orange) are unable to provide good results due to the adverse impact of pixels in shorter exposures. A key limitation of both methods is not explicitly modeling camera noise, which we address in our formulation.}}

\rev{\figref{results-synthetic} depicts results for the two versions of our weighted least-squares solution (\eqref{wls}), showing both pairwise connectivity (in red) and the greedy \ac{MST} heuristic (in purple). Both our solutions use weights that model the noise characteristics of the camera. Thus, they result in better estimates with smaller errors. Overall, the greedy \ac{MST} heuristic is the best-performing estimator of exposure ratios.}

% \begin{table}
% \centering
% \caption{The execution times and properties of the different estimation methods whose accuracies are plotted in \figref{results-synthetic}}
% \label{tab:execution-times}
% \begin{tabular}{@{}cccc@{}}
% \toprule
% & BTF {[}3{]} & MST $\hat{e}_\textrm{WLS}$ & Tiled $\hat{e}_\textrm{WLS}$ \\ \midrule
% Weights & Histogram counts & Noise model & Noise model\\
% Average time (s) & 11.69 & 0.381 & 0.392 \\ \bottomrule
% \end{tabular}
% \end{table}

\textbf{Execution times:}
\tci{Histogram matching used in \cite{cerman2006exposure} is computationally expensive for high-resolution images, resulting in an average execution time of 2.11 seconds. Further, we observed that the expensive iterative procedure for removing outliers is needed for good estimates increasing time to 8.54 seconds. Our tiled reduction (\secref{tiled}) results in a significantly faster average execution time of 0.265 seconds for pairwise connected exposures and 0.29 seconds for the greedy \ac{MST} solution. The reported times were computed on an \emph{Intel i7-8700 CPU} for images of resolution $4312 \times 2868$.}
% \RM{Mention image resolution.}

\subsection{Performance on real captures}

\begin{figure}[t]
    \centering
    % \includegraphics[width=0.32\columnwidth]{images/results/lights/032_2500_50_3100_650_base_annotated.png} \hfill
    % \includegraphics[width=0.32\columnwidth]{images/results/lights/032_2500_50_3100_650_cerman.png} \hfill
    % \includegraphics[width=0.32\columnwidth]{images/results/lights/032_2500_50_3100_650_mst.png} \\ [1ex]
    % \subfloat[EXIF \label{fig:light-base}] {\includegraphics[width=0.32\columnwidth]{images/results/lights/033_2500_600_3300_1400_base_annotated.png}} \hfill
    % \subfloat[BTF~\cite{cerman2006exposure} \label{fig:light-cerman}] {\includegraphics[width=0.32\columnwidth]{images/results/lights/033_2500_600_3300_1400_cerman.png}} \hfill
    % \subfloat[$\bm{\hat{e}_\mathrm{WLS}}$ (Ours) \label{fig:light-mst}] {\includegraphics[width=0.32\columnwidth]{images/results/lights/033_2500_600_3300_1400_mst.png}} \\ [-1ex]
    % % \subfloat[EXIF \label{fig:light-base}]{\includegraphics[width=.32\columnwidth]{images/results/lights/105_2400_500_2700_800_base_annotated.png}} \hfill
    % % \subfloat[BTF~\cite{cerman2006exposure} \label{fig:light-cerman}]{\includegraphics[width=.32\columnwidth]{images/results/lights/105_2400_500_2700_800_cerman.png}} \hfill
    % % \subfloat[$\bm{\hat{e}_\mathrm{WLS}}$ (Ours) \label{fig:light-mst}]{\includegraphics[width=.32\columnwidth]{images/results/lights/105_2400_500_2700_800_mst.png}}
    \includegraphics[width=\columnwidth]{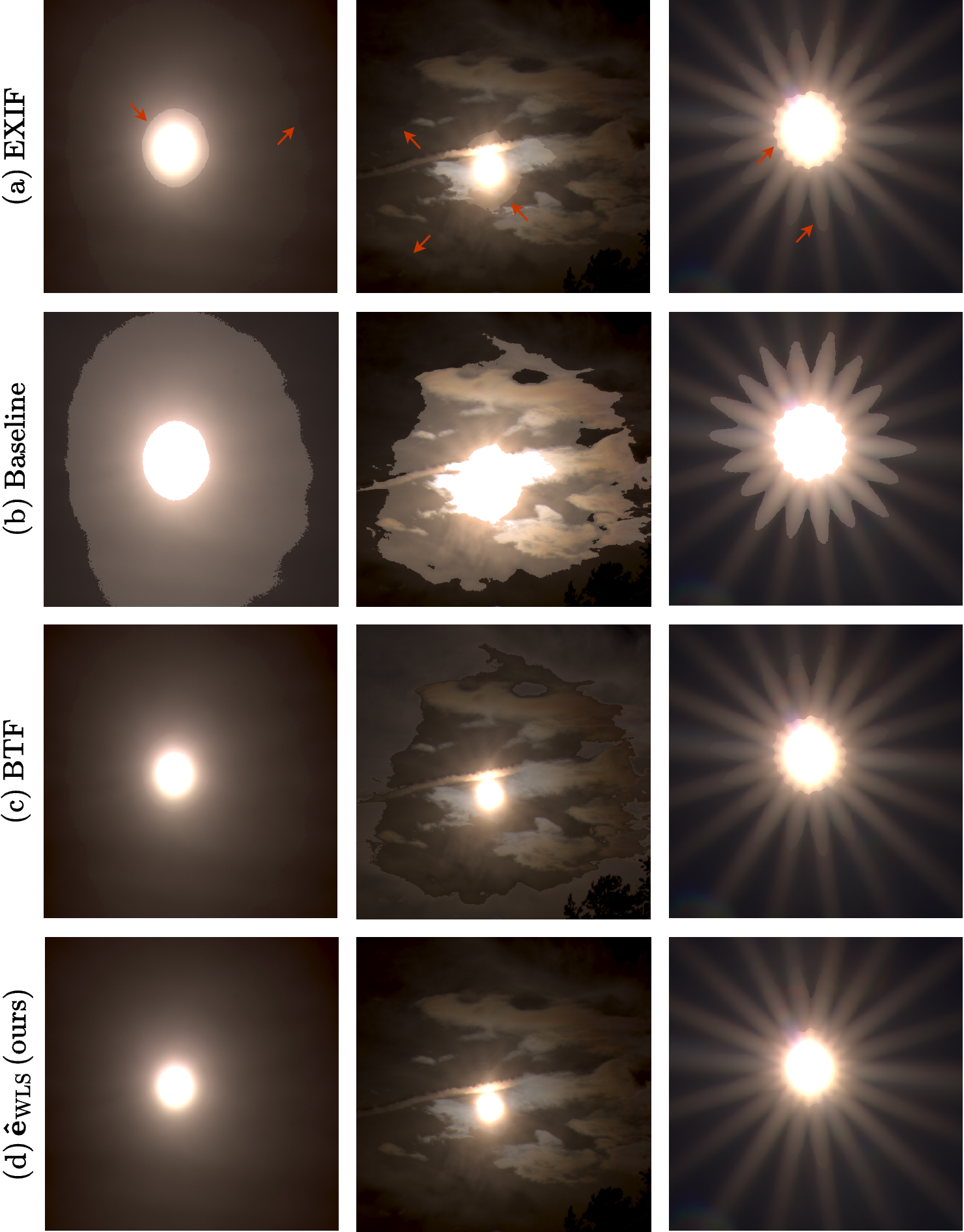}
    \caption{Zoomed in patches of light sources, appropriately exposed to highlight banding close to the source due to inaccurately reported \ac{EXIF} exposures (first row). \tcirev{The baseline method (second row) that does not model camera noise is unable to estimate shorter exposures, resulting in artifact-ridden reconstructions.} Similarly, only some exposures can be aligned with histogram weights based on \ac{BTF} (third row). By modeling camera noise, we simultaneously align all exposures (last row).}
    \label{fig:results-light}
\end{figure}

Next, we show the advantage of our proposed exposure estimation over naive merging using \ac{EXIF} data and compare it with the histogram-based \ac{BTF} approach \cite{cerman2006exposure}.
% We carefully selected image stacks from the SI-HDR dataset~\cite{hanji2022sihdr} that do not contain camera or scene motion. This let us skip the outlier removal (\secref{outlier}), which was validated in \cite{cerman2006exposure}.
\rev{The RAW images have a bit-depth of 14 and are linearly related to the scene radiance. We did not apply \ac{CRF} or tone-mapping for any image shown in this section. Thus, all artifacts visible are due to incorrect exposures.} For all results in this section, we use calibration-free weights given by \eqref{weights-prime}.

Banding artifacts are visible for scenes containing a smooth gradient at the transition point when one of the input exposures saturates. This frequently occurs very close to bright light sources such as the Sun during the day or street lights at night, as indicated by red arrows in \figref{results-light}~(a). Pixels close to light sources tend to be unsaturated only in shorter exposures where other parts of the image are strongly affected by noise. \tcirev{Aligning exposures under these conditions is challenging, resulting in the poor performance of the baseline as well as the histogram-based \ac{BTF} weights~\cite{cerman2006exposure}. The baseline method obtained by directly solving \eqref{expectation-ratio} completely fails to recover exposure ratios.} When using the \ac{BTF} weights, the banding artifacts also persist at the same locations or are introduced in other locations as shown in \figref{results-light}~(c). By accounting for heteroskedastic noise with inverse-variance weights in \figref{results-light}~(d), we can correctly align all the exposures to produce banding-free results.

\begin{figure*}
    \centering
    \includegraphics[width=\textwidth]{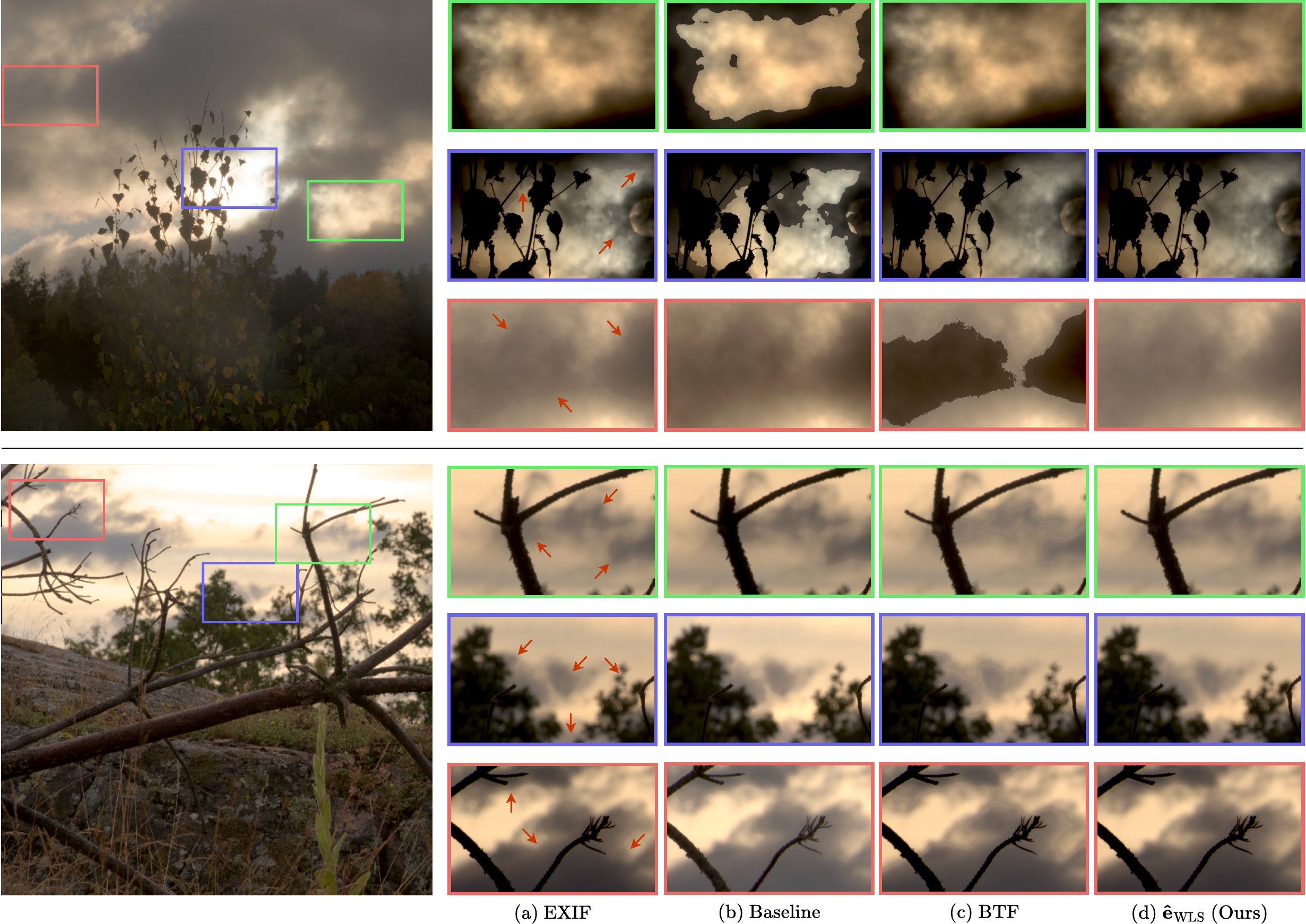}
    \caption{Fixing banding in smooth image gradients when images contain sufficient well-exposed pixels to populate the linear system. Our inverse-variance solution (last column) is more robust and produces better reconstructions.}
    \label{fig:sky-blur}
\end{figure*}

Even in the absence of point light sources, banding can appear at natural smooth gradients or due to defocus blur, as shown in \figref{sky-blur}. Large regions of the images are well-exposed in such scenes, and most methods work reasonably well. \tcirev{However, the baseline and \ac{BTF} weights can still sometimes fail to recover the exposure ratio for shorter exposures (for example, see the red patch in the third column and green and blue patches in the second column of the first scene)}. Our noise-model motivated approach consistently aligns all exposures to produce banding-free reconstructions across the scenes.

\subsection{Deghosting for scenes with motion}

\begin{figure}
    \centering
    \includegraphics[width=0.32\columnwidth]{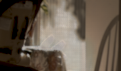} \hfill
    \includegraphics[width=0.32\columnwidth]{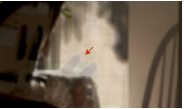} \hfill
    \includegraphics[width=0.32\columnwidth]{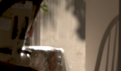} \\
    \subfloat[Aligned \\ exposures \label{fig:deghosting-aligned}]{\includegraphics[width=0.32\columnwidth]{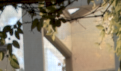}} \hfill
    \subfloat[Corrupted \\ exposures \label{fig:deghosting-corrupted}]{\includegraphics[width=0.32\columnwidth]{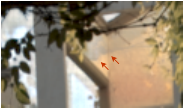}} \hfill
    \subfloat[Reference \label{fig:deghosting-ref}]{\includegraphics[width=0.32\columnwidth]{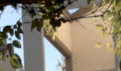}}
    \caption{\tci{Comparison of HDR reconstructions of an attention-based deep network~\cite{yan2020anl} with aligned (first column) and corrupted (middle column) exposures. These insets are from the HDR deghosting dataset~\cite{kalantari2017deep}, which the deghosting network was trained on.}}
    \label{fig:deghosting}
\end{figure}

\tci{When multi-image stacks with misaligned exposures are fused, the resulting HDR reconstructions often exhibit accentuated ghosting artifacts. Even recent deep learning-based methods~\cite{kalantari2017deep,yan2020nl,yan2020anl} are unable to account for inaccurate exposure ratios. \figref{deghosting-corrupted} illustrates the artifacts that arise when exposures are corrupted according to \eqref{res-error}. It is important to note that these artifacts are in addition to the ghosting artifacts visible in \figref{deghosting-aligned}. Our proposed algorithm, which includes tile-based outlier detection to account for scene motion, effectively aligns exposures.}

\subsection{Measuring display MTF}
\label{sec:display-mtf}
\begin{figure}
    \centering
    \includegraphics[width=\columnwidth]{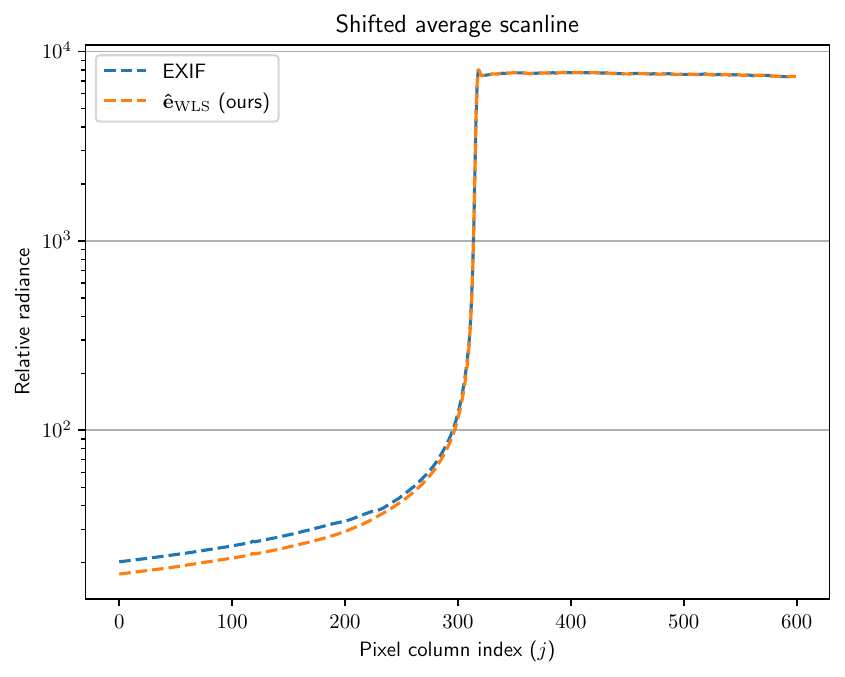}
    \caption{Edge profile of an \ac{HDR} display recovered from a slant edge captured with an exposure stack. Errors in \ac{EXIF} metadata lead to deviation from the actual edge spread function, affecting MTF estimation. On inspecting the captured images, we observed the banding artifacts close to the captured slant-edge when the image was merged with EXIF exposures.
    % \RM{Luminance (or radiance) is typically reported on the log-10 scale: 1, 10, 100.... Because camera does not measure luminance and luminance is not needed for the MTF measurements, label the axis "relative radiance".}
    }
    \label{fig:scanline-sfr}
\end{figure}

Digital cameras are often used as inexpensive light measuring instruments, for example, for measuring the spatial characteristic of an electronic display, such as spatial uniformity of a modulation transfer function (MTF)~\cite{zhang2012mtf}. Inaccuracies in pixel values due to exposure misalignment may lead to errors in the estimated display characteristics. Here, we demonstrate the need for exposure alignment for measuring MTF of HDR displays.

The MTF measurement involves taking an image of a slanted edge~\cite{samei1998slantedge}, where one side has its pixels set to 0 and the other side to the maximum pixel value. Modern displays can reach very high contrast levels, so such a slanted-edge image can be faithfully captured only using an exposure stack. The inaccuracies of exposure times can easily introduce bias to our measurements. 

\figref{scanline-sfr} shows the edge profile computed with (blue) and without (orange) exposure estimation. The difference in reconstructions is due to inaccurate EXIF exposure values, visible as banding in the merged image. This translates to the bias at low-radiance pixels in \figref{scanline-sfr} and shows that errors in exposure times can easily lead to biases in the final MTF estimates.

% \RM{What is the difference between SFR and MTF? I believe MTF is more frequently used.}
%Some applications, such as display uniformity calibration and estimation of the \ac{SFR}~\cite{iso2000sfr}\RM{What is SFR? You should explain the acronym the first time you use it.}, require exact correspondence between physical quantities reconstructed pixel intensities. Imprecisely scaled pixels due to incorrect \ac{EXIF} metadata will adversely affect such applications.

%While a spectrophotometer will provide accurate luminance readings of bright stimuli, it is unsuitable for measuring dark surfaces. Moreover, taking multiple readings at different spatial locations is a tedious, time-consuming process, and the preferred approach is to use a high-quality camera. To estimate the \ac{SFR} of an \ac{HDR} display, we captured an exposure stack of slant-edge. \figref{scanline-sfr} shows the edge profile computed with (blue) and without (orange) exposure estimation. The difference in reconstructions at the transition is due to misaligned exposure values, visible as banding in the images. This is also the reason for the small bias for low-luminance pixels (zoom in at $j < 200$). It is thus imperative to first align exposures for accurate \ac{SFR} estimation.
\section{Conclusions}
We propose to estimate relative exposures directly from images of an \ac{HDR} exposure stack instead of relying on inaccurate camera EXIF metadata. The problem can be formulated as a weighted linear system of equations with Tikhonov regularization. Our formulation considers the noise characteristics of a camera to derive weights. We show that good performance on a large number of scenes is possible without the need for camera- and gain-specific noise parameters.

We also describe an efficient approach based on the union of the highest weighted spanning trees of the exposure multigraph to reduce the size of the system for large images. When applied to multi-image \ac{HDR} stacks, our method eliminates banding artifacts at smooth image gradients close to light sources or due to defocus blur. The exposure-aligned images are closer to physical quantities, making our work essential for using a camera as a light-measurement instrument. 

% \pagebreak

\section*{Acknowledgments}

This project has received funding from the European Research Council (ERC) under the European Union's Horizon 2020 research and innovation programme (grant agreement N$^\circ$ 725253--EyeCode).

\appendix[Moments of a transformed random variable]
Here, we use the Taylor expansion to derive expressions for the first and second moments (expected value and variance respectively) of a random variable under an arbitrary invertible function. In \secref{heteroskedastic} we apply these results to obtain expressions for the $\log$ of a normally distributed random variable.

% \subsection{Moments of a transformed random variable}
% \RM{If the method comes from a textbook, add a reference.}
To get an approximate variance for a nonlinear function of a random variable, we describe the Delta method~\cite{oehlert1992delta,hoef2012delta} that uses a Taylor expansion. Let $Z$ be an asymptotically normal random variable with a known mean $\mu_Z$ and variance $\sigma_Z^2$, and let $f$ be an invertible and differentiable function. To get expressions for the expected value and variance of $f(Z)$ we use its first order Taylor expansion about the mean $\mu_Z$,
% \RM{Do you need to make any assumption, such as that $\mu_Z$ is normally distributed?}
\begin{equation}
\begin{split}
    f(Z) = \,& f(\mu_Z) + f'(\mu_Z) (Z - \mu_Z) \\
    &+ \textrm{higher-order terms}\,.
\end{split}
\end{equation}
Ignoring the diminishing contributions of higher-order terms, the expected value is,
\begin{equation} \label{eq:appendix-ev}
\begin{split}
    \E[f(Z)] &\approx \E[f(\mu_Z)] + \E[f'(\mu_Z) (Z - \mu_Z)] \\
    &= \E[f(\mu_Z)] + f'(\mu_Z) (\E[Z] - \E[\mu_Z)]) \\
    &= f(\mu_Z) + f'(\mu_Z) (\mu_Z - \mu_Z) \\
    &= f(\mu_Z)\,,
\end{split}
\end{equation}
% \RM{Please explain how the expected values can be reduced to $f(\mu_Z)$. It is not clear to me.}
and the variance is
\begin{equation} \label{eq:appendix-var}
\begin{split}
    \V[f(Z)] &\approx \V[f(\mu_Z)] + \V[f'(\mu_Z) (Z - \mu_Z)] \\
    &= \V[f(\mu_Z)] + f'(\mu_Z)^2 \, \V[Z - \mu_Z] \\
    &= \V[f(\mu_Z)] + f'(\mu_Z)^2 \, (\V[Z] + \V[\mu_Z]) \\
    &= 0 + f'(\mu_Z)^2 \, (\sigma_Z^2 + 0) \\
    &= f'(\mu_Z)^2 \, \sigma_Z^2\,.
\end{split}
\end{equation}
% \RM{Please add more explanation, I cannot follow the steps.}
% \PH{Is this better?}

\bibliographystyle{IEEEtranS}
\bibliography{references}

% Generated by IEEEtranS.bst, version: 1.14 (2015/08/26)
\begin{thebibliography}{10}
\providecommand{\url}[1]{#1}
\csname url@samestyle\endcsname
\providecommand{\newblock}{\relax}
\providecommand{\bibinfo}[2]{#2}
\providecommand{\BIBentrySTDinterwordspacing}{\spaceskip=0pt\relax}
\providecommand{\BIBentryALTinterwordstretchfactor}{4}
\providecommand{\BIBentryALTinterwordspacing}{\spaceskip=\fontdimen2\font plus
\BIBentryALTinterwordstretchfactor\fontdimen3\font minus
  \fontdimen4\font\relax}
\providecommand{\BIBforeignlanguage}[2]{{%
\expandafter\ifx\csname l@#1\endcsname\relax
\typeout{** WARNING: IEEEtranS.bst: No hyphenation pattern has been}%
\typeout{** loaded for the language `#1'. Using the pattern for}%
\typeout{** the default language instead.}%
\else
\language=\csname l@#1\endcsname
\fi
#2}}
\providecommand{\BIBdecl}{\relax}
\BIBdecl

\bibitem{abdelhamed2019noiseflow}
A.~Abdelhamed, M.~A. Brubaker, and M.~S. Brown, ``{Noise Flow: Noise Modeling
  with Conditional Normalizing Flows},'' in \emph{International Conference on
  Computer Vision (ICCV)}, 2019.

\bibitem{aguerrebere2012study}
\BIBentryALTinterwordspacing
C.~Aguerrebere, J.~Delon, Y.~Gousseau, and P.~Mus{\'e}, ``{Study of the digital
  camera acquisition process and statistical modeling of the sensor raw
  data},'' Laboratoire Traitement et Communication de l'Information, Instituto
  de Ingeniería Eléctrica, Tech. Rep., Sep. 2012. [Online]. Available:
  \url{https://hal.archives-ouvertes.fr/hal-00733538}
\BIBentrySTDinterwordspacing

\bibitem{cerman2006exposure}
L.~Cerman and V.~Hlavac, ``Exposure time estimation for high dynamic range
  imaging with hand held camera,'' in \emph{Proc. of Computer Vision Winter
  Workshop, Czech Republic}.\hskip 1em plus 0.5em minus 0.4em\relax Citeseer,
  2006.

\bibitem{chang2020learning}
K.-C. Chang, R.~Wang, H.-J. Lin, Y.-L. Liu, C.-P. Chen, Y.-L. Chang, and H.-T.
  Chen, ``Learning camera-aware noise models,'' in \emph{Proceedings of
  European Conference on Computer Vision (ECCV)}, 2020.

\bibitem{chen2021attention}
J.~Chen, Z.~Yang, T.~N. Chan, H.~Li, J.~Hou, and L.-P. Chau, ``Attention-guided
  progressive neural texture fusion for high dynamic range image restoration,''
  \emph{IEEE Transactions on Image Processing}, vol.~31, pp. 2661--2672, 2022.

\bibitem{chen2020learning}
Y.~Chen, G.~Jiang, M.~Yu, Y.~Yang, and Y.-S. Ho, ``Learning stereo high dynamic
  range imaging from a pair of cameras with different exposure parameters,''
  \emph{IEEE Transactions on Computational Imaging}, vol.~6, pp. 1044--1058,
  2020.

\bibitem{debevec1997recovering}
\BIBentryALTinterwordspacing
P.~E. Debevec and J.~Malik, ``Recovering high dynamic range radiance maps from
  photographs,'' in \emph{Proceedings of the 24th Annual Conference on Computer
  Graphics and Interactive Techniques}, ser. SIGGRAPH '97.\hskip 1em plus 0.5em
  minus 0.4em\relax USA: ACM Press/Addison-Wesley Publishing Co., 1997, p.
  369–378. [Online]. Available: \url{https://doi.org/10.1145/258734.258884}
\BIBentrySTDinterwordspacing

\bibitem{eilertson2017hdr}
G.~Eilertsen, J.~Kronander, G.~Denes, R.~Mantiuk, and J.~Unger, ``Hdr image
  reconstruction from a single exposure using deep cnns,'' \emph{ACM
  Transactions on Graphics (TOG)}, vol.~36, no.~6, 2017.

\bibitem{eppstein1990finding}
D.~Eppstein, ``Finding the k smallest spanning trees,'' in \emph{Scandinavian
  Workshop on Algorithm Theory}.\hskip 1em plus 0.5em minus 0.4em\relax
  Springer, 1990, pp. 38--47.

\bibitem{fairchild2007hdr}
M.~D. Fairchild, ``The hdr photographic survey,'' in \emph{Color and imaging
  conference}, vol. 2007.\hskip 1em plus 0.5em minus 0.4em\relax Society for
  Imaging Science and Technology, 2007, pp. 233--238.

\bibitem{foi2009heteroskedastic}
\BIBentryALTinterwordspacing
A.~Foi, ``Clipped noisy images: Heteroskedastic modeling and practical
  denoising,'' \emph{Signal Processing}, vol.~89, no.~12, pp. 2609--2629, 2009,
  special Section: Visual Information Analysis for Security. [Online].
  Available:
  \url{https://www.sciencedirect.com/science/article/pii/S0165168409001996}
\BIBentrySTDinterwordspacing

\bibitem{foi2008noise}
A.~Foi, M.~Trimeche, V.~Katkovnik, and K.~Egiazarian, ``Practical
  poissonian-gaussian noise modeling and fitting for single-image raw-data,''
  \emph{IEEE Transactions on Image Processing}, vol.~17, no.~10, pp.
  1737--1754, 2008.

\bibitem{Gallo2016}
O.~Gallo and P.~Sen, ``{Stack-Based Algorithms for HDR Capture and
  Reconstruction},'' in \emph{High Dynamic Range Video}.\hskip 1em plus 0.5em
  minus 0.4em\relax Elsevier, 2016, pp. 85--119.

\bibitem{granados2010optimal}
M.~Granados, B.~Ajdin, M.~Wand, C.~Theobalt, H.-P. Seidel, and H.~P. Lensch,
  ``Optimal hdr reconstruction with linear digital cameras,'' in \emph{2010
  IEEE Computer Society Conference on Computer Vision and Pattern
  Recognition}.\hskip 1em plus 0.5em minus 0.4em\relax IEEE, 2010, pp.
  215--222.

\bibitem{grossberg2002brightness}
M.~D. Grossberg and S.~K. Nayar, ``What can be known about the radiometric
  response from images?'' in \emph{Computer Vision --- ECCV 2002}, A.~Heyden,
  G.~Sparr, M.~Nielsen, and P.~Johansen, Eds.\hskip 1em plus 0.5em minus
  0.4em\relax Berlin, Heidelberg: Springer Berlin Heidelberg, 2002, pp.
  189--205.

\bibitem{hajisharif2015adaptive}
S.~Hajisharif, J.~Kronander, and J.~Unger, ``Adaptive dualiso hdr
  reconstruction,'' \emph{EURASIP Journal on Image and Video Processing}, vol.
  2015, no.~1, p.~41, 2015.

\bibitem{hanji2021hdr4cv}
\BIBentryALTinterwordspacing
P.~Hanji, M.~Z. Alam, N.~Giuliani, H.~Chen, and R.~K. Mantiuk, ``Hdr4cv: High
  dynamic range dataset with adversarial illumination for testing computer
  vision methods,'' \emph{Journal of Imaging Science and Technology}, 2021.
  [Online]. Available:
  \url{http://www.cl.cam.ac.uk/research/rainbow/projects/hdr4cv-dataset/}
\BIBentrySTDinterwordspacing

\bibitem{hanji2022sihdr}
\BIBentryALTinterwordspacing
P.~Hanji, R.~K. Mantiuk, G.~Eilertsen, S.~Hajisharif, and J.~Unger,
  ``Comparison of single image hdr reconstruction methods — the caveats of
  quality assessment,'' in \emph{Special Interest Group on Computer Graphics
  and Interactive Techniques Conference Proceedings (SIGGRAPH '22 Conference
  Proceedings)}, 2022. [Online]. Available:
  \url{https://www.cl.cam.ac.uk/research/rainbow/projects/sihdr_benchmark/}
\BIBentrySTDinterwordspacing

\bibitem{hanji2020noise}
\BIBentryALTinterwordspacing
P.~Hanji, F.~Zhong, and R.~K. Mantiuk, ``Noise-aware merging of high dynamic
  range image stacks without camera calibration,'' in \emph{Advances in Image
  Manipulation (ECCV workshop)}.\hskip 1em plus 0.5em minus 0.4em\relax
  Springer, 2020, pp. 376--391. [Online]. Available:
  \url{https://doi.org/10.1007/978-3-030-67070-2_23}
\BIBentrySTDinterwordspacing

\bibitem{hasinoff2016burst}
S.~W. Hasinoff, D.~Sharlet, R.~Geiss, A.~Adams, J.~T. Barron, F.~Kainz,
  J.~Chen, and M.~Levoy, ``Burst photography for high dynamic range and
  low-light imaging on mobile cameras,'' \emph{ACM Transactions on Graphics
  (TOG)}, vol.~35, no.~6, p. 192, 2016.

\bibitem{hasinoff2010noise}
S.~Hasinoff, F.~Durand, and W.~Freeman, ``{Noise-optimal capture for high
  dynamic range photography},'' in \emph{CVPR}.\hskip 1em plus 0.5em minus
  0.4em\relax IEEE, 2010, pp. 553--560.

\bibitem{heide2014flexisp}
F.~Heide, M.~Steinberger, Y.-T. Tsai, M.~Rouf, D.~Pajak, D.~Reddy, O.~Gallo,
  J.~Liu, W.~Heidrich, K.~Egiazarian \emph{et~al.}, ``Flexisp: A flexible
  camera image processing framework,'' \emph{ACM Transactions on Graphics
  (TOG)}, vol.~33, no.~6, pp. 1--13, 2014.

\bibitem{hoef2012delta}
\BIBentryALTinterwordspacing
J.~M.~V. Hoef, ``Who invented the delta method?'' \emph{The American
  Statistician}, vol.~66, no.~2, pp. 124--127, 2012. [Online]. Available:
  \url{https://doi.org/10.1080/00031305.2012.687494}
\BIBentrySTDinterwordspacing

\bibitem{kalantari2017deep}
N.~K. Kalantari and R.~Ramamoorthi, ``Deep high dynamic range imaging of
  dynamic scenes,'' \emph{ACM Transactions on Graphics (Proceedings of SIGGRAPH
  2017)}, vol.~36, no.~4, 2017.

\bibitem{Karaduzovic-Hadziabdic2016}
K.~Karaduzovic-Hadziabdic, J.~H. Telalovic, and R.~Mantiuk, ``{Subjective and
  Objective Evaluation of Multi-exposure High Dynamic Range Image Deghosting
  Methods},'' in \emph{Eurographics 2016 - Short Papers}, 2016.

\bibitem{katoh1981algorithm}
N.~Katoh, T.~Ibaraki, and H.~Mine, ``An algorithm for finding k minimum
  spanning trees,'' \emph{SIAM Journal on Computing}, vol.~10, no.~2, pp.
  247--255, 1981.

\bibitem{Kim2020}
\BIBentryALTinterwordspacing
M.~Kim, M.~Azimi, and R.~K. Mantiuk, ``{Perceptually motivated model for
  predicting banding artefacts in high-dynamic range images},'' \emph{Color and
  Imaging Conference}, vol. 2020, no.~28, pp. 42--48, nov 2020. [Online].
  Available:
  \url{https://www.ingentaconnect.com/content/10.2352/issn.2169-2629.2020.28.8}
\BIBentrySTDinterwordspacing

\bibitem{konnik2014high}
M.~Konnik and J.~Welsh, ``High-level numerical simulations of noise in ccd and
  cmos photosensors: review and tutorial,'' \emph{arXiv preprint
  arXiv:1412.4031}, 2014.

\bibitem{ma2017mefssim}
K.~Ma, Z.~Duanmu, H.~Yeganeh, and Z.~Wang, ``Multi-exposure image fusion by
  optimizing a structural similarity index,'' \emph{IEEE Transactions on
  Computational Imaging}, vol.~4, no.~1, pp. 60--72, 2018.

\bibitem{mitsunaga1999radiometric}
T.~Mitsunaga and S.~Nayar, ``Radiometric self calibration,'' in
  \emph{Proceedings. 1999 IEEE Computer Society Conference on Computer Vision
  and Pattern Recognition (Cat. No PR00149)}, vol.~1, 1999, pp. 374--380 Vol.
  1.

\bibitem{oehlert1992delta}
\BIBentryALTinterwordspacing
G.~W. Oehlert, ``A note on the delta method,'' \emph{The American
  Statistician}, vol.~46, no.~1, pp. 27--29, 1992. [Online]. Available:
  \url{https://www.tandfonline.com/doi/abs/10.1080/00031305.1992.10475842}
\BIBentrySTDinterwordspacing

\bibitem{Perez-Pellitero_2021_CVPR}
E.~Perez-Pellitero, S.~Catley-Chandar, A.~Leonardis, and R.~Timofte, ``Ntire
  2021 challenge on high dynamic range imaging: Dataset, methods and results,''
  in \emph{Proceedings of the IEEE/CVF Conference on Computer Vision and
  Pattern Recognition (CVPR) Workshops}, June 2021, pp. 691--700.

\bibitem{prabhakar2021self}
K.~R. Prabhakar, S.~Agrawal, and R.~V. Babu, ``Self-gated memory recurrent
  network for efficient scalable hdr deghosting,'' \emph{IEEE Transactions on
  Computational Imaging}, vol.~7, pp. 1228--1239, 2021.

\bibitem{Pu2020robust}
Z.~Pu, P.~Guo, M.~S. Asif, and Z.~Ma, ``Robust high dynamic range (hdr) imaging
  with complex motion and parallax,'' in \emph{Proceedings of the Asian
  Conference on Computer Vision (ACCV)}, November 2020.

\bibitem{Rodriguez2019}
\BIBentryALTinterwordspacing
R.~G. Rodríguez, J.~Vazquez-Corral, and M.~Bertalmío, ``Issues with common
  assumptions about the camera pipeline and their impact in hdr imaging from
  multiple exposures,'' \emph{https://doi.org/10.1137/19M1250248}, vol.~12, pp.
  1627--1642, 10 2019. [Online]. Available:
  \url{https://epubs.siam.org/doi/10.1137/19M1250248}
\BIBentrySTDinterwordspacing

\bibitem{samei1998slantedge}
E.~Samei, M.~J. Flynn, and D.~A. Reimann, ``A method for measuring the
  presampled mtf of digital radiographic systems using an edge test device,''
  \emph{Medical physics}, vol.~25, no.~1, pp. 102--113, 1998.

\bibitem{Marcel2020LDRHDR}
M.~S. Santos, R.~Tsang, and N.~Khademi~Kalantari, ``Single image hdr
  reconstruction using a cnn with masked features and perceptual loss,''
  \emph{ACM Transactions on Graphics}, vol.~39, no.~4, 7 2020.

\bibitem{tani1995photographic}
T.~Tani, \emph{Photographic sensitivity: theory and mechanisms}.\hskip 1em plus
  0.5em minus 0.4em\relax Oxford University Press on Demand, 1995, no.~8.

\bibitem{Tomaszewska2007homography}
A.~Tomaszewska and R.~Mantiuk, ``Image registration for multi-exposure high
  dynamic range image acquisition,'' in \emph{Proceedings of the International
  Conference in Central Europe on Computer Graphics, Visualization and Computer
  Vision}, 01 2007.

\bibitem{tu2020banding}
Z.~Tu, J.~Lin, Y.~Wang, B.~Adsumilli, and A.~C. Bovik, ``Bband index: A
  no-reference banding artifact predictor,'' in \emph{ICASSP 2020 - 2020 IEEE
  International Conference on Acoustics, Speech and Signal Processing
  (ICASSP)}, 2020, pp. 2712--2716.

\bibitem{tursun2015state}
O.~T. Tursun, A.~O. Aky{\"u}z, A.~Erdem, and E.~Erdem, ``The state of the art
  in hdr deghosting: a survey and evaluation,'' in \emph{Computer Graphics
  Forum}, vol.~34, no.~2.\hskip 1em plus 0.5em minus 0.4em\relax Wiley Online
  Library, 2015, pp. 683--707.

\bibitem{yan2020anl}
\BIBentryALTinterwordspacing
T.~V.V, ``High dynamic range image synthesis via attention non-local network,''
  2020. [Online]. Available: \url{github.com/tuvovan/ANL-HDRI}
\BIBentrySTDinterwordspacing

\bibitem{wang2021deep}
L.~Wang and K.-J. Yoon, ``Deep learning for hdr imaging: State-of-the-art and
  future trends,'' \emph{IEEE Transactions on Pattern Analysis and Machine
  Intelligence}, 2021.

\bibitem{wang2016banding}
Y.~Wang, S.-U. Kum, C.~Chen, and A.~Kokaram, ``A perceptual visibility metric
  for banding artifacts,'' in \emph{2016 IEEE International Conference on Image
  Processing (ICIP)}, 2016, pp. 2067--2071.

\bibitem{yan2020nl}
Q.~Yan, L.~Zhang, Y.~Liu, Y.~Zhu, J.~Sun, Q.~Shi, and Y.~Zhang, ``Deep hdr
  imaging via a non-local network,'' \emph{IEEE Transactions on Image
  Processing}, vol.~29, pp. 4308--4322, 2020.

\bibitem{yang2018image}
X.~Yang, K.~Xu, Y.~Song, Q.~Zhang, X.~Wei, and R.~W. Lau, ``Image correction
  via deep reciprocating hdr transformation,'' in \emph{Proceedings of the IEEE
  Conference on Computer Vision and Pattern Recognition}, 2018, pp. 1798--1807.

\bibitem{ye2021psfn}
\BIBentryALTinterwordspacing
Q.~Ye, J.~Xiao, K.-m. Lam, and T.~Okatani, \emph{Progressive and Selective
  Fusion Network for High Dynamic Range Imaging}.\hskip 1em plus 0.5em minus
  0.4em\relax New York, NY, USA: Association for Computing Machinery, 2021, p.
  5290–5297. [Online]. Available:
  \url{https://doi.org/10.1145/3474085.3475651}
\BIBentrySTDinterwordspacing

\bibitem{zhang2012mtf}
X.~Zhang, T.~Kashti, D.~Kella, T.~Frank, D.~Shaked, R.~Ulichney, M.~Fischer,
  and J.~P. Allebach, ``Measuring the modulation transfer function of image
  capture devices: what do the numbers really mean?'' in \emph{Image Quality
  and System Performance IX}, vol. 8293.\hskip 1em plus 0.5em minus 0.4em\relax
  SPIE, 2012, pp. 64--74.

\end{thebibliography}

\end{document}